\documentclass{article}

\PassOptionsToPackage{sort&compress,numbers}{natbib}
\usepackage[preprint]{neurips_2024}

\usepackage[utf8]{inputenc} 
\usepackage[T1]{fontenc}    
\usepackage{hyperref}       
\usepackage{url}            
\usepackage{booktabs}       
\usepackage{amsfonts}       
\usepackage{nicefrac}       
\usepackage{microtype}      
\usepackage{xcolor}         

\usepackage{graphicx}
\usepackage{amsmath,amssymb}
\usepackage{adjustbox}
\usepackage{algorithm}
\usepackage[noend]{algpseudocode}

\newcommand{\be}{\begin{eqnarray}} 
\newcommand{\ee}{\end{eqnarray}}
\newcommand{\D}{\mathrm{d}}
\renewcommand{\vec}{\mathbf}

\newcommand{\bxi}{\boldsymbol\xi}
\newcommand{\blam}{\boldsymbol\Lambda}
\newcommand{\btheta}{\boldsymbol\theta}
\newcommand{\bsigma}{\boldsymbol\Sigma}
\newcommand{\bmu}{\boldsymbol\mu}

\title{Generative modelling with jump-diffusions}

\author{%
  Adrian Baule\\
  School of Mathematical Sciences\\
  Queen Mary University of London\\
  London E1 4NS, United Kingdom \\
  \texttt{a.baule@qmul.ac.uk} \\
 }

\begin{document}

\maketitle

\begin{abstract}
Score-based diffusion models generate samples from an unknown target distribution using a time-reversed diffusion process. While such models represent state-of-the-art approaches in industrial applications such as artificial image generation, it has recently been noted that their performance can be further improved by considering injection noise with heavy tailed characteristics. Here, I present a generalization of generative diffusion processes to a wide class of non-Gaussian noise processes. I consider forward processes driven by standard Gaussian noise with super-imposed Poisson jumps representing a finite activity L\'evy process. The generative process is shown to be governed by a {\em generalized score function} that depends on the jump amplitude distribution {and can be estimated by minimizing a simple MSE loss as in conventional Gaussian models}. Both probability flow ODE and SDE formulations are derived using basic technical effort. A detailed implementation for a pure jump process with Laplace distributed amplitudes yields a generalized score function in closed analytical form and is shown to outperform the equivalent Gaussian model in specific parameter regimes.
\end{abstract}

\section{Introduction}

Generative diffusion models transform a target distribution into a simple distribution by successively adding noise and then reversing this process. By sampling from the simple distribution, the reversed (or denoising) process is thus able to generate a new target sample. Starting from this seminal concept \cite{Sohl-Dickstein:2015aa}, two main implementations of diffusion models (DMs) have been developed, where, roughly speaking, the denoising process is either formulated in discrete (time) steps and represented by a Markov chain, as implemented for example in denoising score-matching \cite{Vincent:2011aa} and denoising diffusion probabilistic models (DDPMs) \cite{Sohl-Dickstein:2015aa,Ho:2020aa}, or implemented in the form of an ordinary differential equation (ODE) or stochastic differential equation (SDE) in continuous time \cite{Song:2021aa}. {DMs based on ODEs are closely related to optimal transport problems and have been generalized in frameworks known as flow matching \cite{Lipman:2022aa,Liu:2022aa,Tong:2023aa} and stochastic interpolants \cite{Albergo:2023aa}.}

{The overwhelming majority of research on DMs is based on underlying noise processes or distributions with Gaussian characteristics, which facilitates both theoretical treatments and computational implementations.} However, from a performance perspective, it is not a priori obvious that Gaussian statistics represents the optimal choice in specific applications. Indeed, the question of how the performance of generative diffusion models can be improved by changing the characteristics of the underlying noise has recently attracted interest in the research community. In \cite{Nachmani:2021aa} it has been shown that using noise from a Gamma distribution in DDPM improves results for image and speech generation. In the SDE approach of \cite{Nobis:2024aa}, Gaussian noise has been replaced by noise from fractional Brownian motion, which achieves greater pixel-wise diversity and enhanced image quality. The effect of $\alpha$-stable noise has been investigated both in the SDE formulation \cite{Yoon:2023aa} and DDPM \cite{Shariatian:2025aa}, and has been shown to allow for faster and more diverse sampling. In \cite{Pandey:2024aa} DDPM has been implemented with noise from a heavy-tailed Student-t distribution and shown to outperform standard diffusion models in heavy-tail estimation on high-resolution weather datasets. In \cite{Holderrieth:2025aa} a Markovian jump model based on generator matching has been proposed, which achieves state-of-the-art performance in multi-modal protein generation.

In this work, I present a non-Gaussian generalization of the conventional Gaussian framework of diffusion models that includes discontinuous jumps in the noise process. The contributions of the present work are summarized as follows:
\begin{itemize}
\item A score-based generative framework for jump-diffusions as underlying noise process is presented, generalizing the work of \cite{Song:2021aa}.
{\item The framework is based on a generalized score function $\neq \nabla \log\,p(\vec{x},t)$ that can be estimated by minimizing a simple denoising score matching loss function similar to the standard case \cite{Vincent:2011aa}.} 
\item Representations of the generative process in terms of a probability flow ODE and an SDE with jump-diffusion noise are derived.
\item {A detailed implementation for jump amplitudes drawn from a multivariate Laplace distribution is presented (referred to as {\em JL model}), where both conditional distributions and the generalized score function have closed analytical forms. The training and generative sampling is only slightly more computationally costly than for standard DMs and is shown to outperform an equivalent Gaussian model in specific parameter regimes.}
\end{itemize}

\subsection{Relationship with other works}

{\bf DMs driven by non-Gaussian noise}. The present work is closely related to the work of \cite{Yoon:2023aa}, who first considered the case of jumps in the forward process in the L\'evy-Ito model (LIM). While LIM only focuses on the special case of infinite activity $\alpha$-stable noise with $1<\alpha<2$, I consider finite activity L\'evy noise given as a superposition of Gaussian noise with Poisson jumps drawn from an arbitrary normalizable distribution. The mathematical discussion is then much simplified and provides a pedagogical access to non-Gaussian generalizations of the conventional framework. {Moreover, the LIM SDE neglects an additional noise term with finite variation, which arises from the exact time reversal formalism \cite{Conforti:2022aa,Yoon:2023aa}. As such noise and score-function are not exactly balanced unlike in the JL model presented here. The computational cost of training and sampling the JL model is equivalent to LIM and thus provides an interesting non-Gaussian alternative to consider in specific applications. For the simple tests and neural network architecture considered here the JL model is shown to outperform LIM in a range of parameter regimes.}

DDPMs with specific non-Gaussian noise distributions have been investigated in \cite{Nachmani:2021aa} (gamma), \cite{Shariatian:2025aa} ($\alpha$-stable), \cite{Pandey:2024aa} (Student-t). DSM has been extended to generalized Gaussian noise with heavy tails in \cite{Deasy:2022aa}. The work of \cite{Li:2024aa} considers random walks with non-Gaussian increments that converge to standard diffusion models. All these Markov chain approaches are formally different even in the limit of continuous time, since the non-Gaussian increments act at every time step of the underlying process, which is different from the jump structure of a L\'evy process whose jump times are Poisson distributed and thus intermittent.

In \cite{Campbell:2023aa} a generative diffusion model is proposed that includes Poisson jumps between different dimensional spaces. The jumps only act in terms of creating/destroying dimensions and do not denoise the samples within a given dimension.

{{\bf DMs based on general Markov processes}. Recently, frameworks for DMs have been presented that include jump processes on continuous state spaces as well \cite{Holderrieth:2025aa,Ren:2025aa}. The ``generator-matching" framework of \cite{Holderrieth:2025aa} specifies the backward generative process in terms of the generator of a general Markov process and learns its coefficients using a prescribed probability path, thus generalizing the flow matching approach to general SDEs. Extending ideas formulated in \cite{Benton:2024aa}, the denoising framework of \cite{Ren:2025aa} considers a general Markov process as forward process and estimates the corresponding backward process by minimizing a KL divergence as loss function. However, applying this approach to finite activity L\'evy processes generally leads to backward processes with jump distributions that are inhomogeneous in space and time and thus non-trivial to sample from. Likewise in generator-matching, where the backwards generator depends on the prescribed probability path, but contains space- and/or time-inhomogeneous jump distributions already for very simple paths. On the other hand, the generalized score function derived here is shown to balance the noise in the backward process and can be estimated using a simple MSE loss. Instead of estimating the backward jump distribution as in \cite{Holderrieth:2025aa,Ren:2025aa}, the framework thus only requires the estimation of a drift term, which captures both ODE and SDE representations. As such it represents an elegant and simple generalization of the Gaussian case. }

\section{Generative score-based diffusion models}
\label{Sec:intro}

Generative diffusion models sample from an (unknown) target distribution $p_{\rm data}$ by modelling an inverse time process. Starting point in the SDE formulation is the forward diffusion process
\be
\label{forward}
\dot{\vec{Y}}(t)=\vec{f}(\vec{Y},t)+g(t)\bxi(t)
\ee
for an initial data distribution $\vec{Y}(0)\sim p_{\rm data}(\vec{x})$ with sample space $\vec{x}\in \mathbb{R}^d$. The noise is typically taken as Gaussian white noise $\boldsymbol\xi(t)=\boldsymbol\xi^{\rm G}(t)$ with covariance $<\xi^{\rm G}_i(t)\xi^{\rm G}_j(t')>=D\delta_{ij}\delta(t-t')$. The forward process is distributed as $\vec{Y}(t)\sim p(\vec{x},t)$, with the associated probability density function (PDF) $p(\vec{x},t)=\int\D \vec{x}'p(\vec{x},t|\vec{x}')p_{\rm data}(\vec{x})$, where $p(\vec{x},t|\vec{x}')$ denotes the conditional PDF of the forward process for the initial condition $\vec{Y}(0) = \vec{x}'$.

The generative process then starts from  $\vec{X}(0)\sim p(\vec{x},T)$ and generates samples of the unknown $p_{\rm data}$ using time-reversal. The time-reversed process is given by \cite{Anderson:1982aa}
\be
\label{reverse}
\dot{\vec{X}}(t)=-\vec{f}(\vec{X},T-t)+Dg^2(T-t)\nabla\log\,p(\vec{X},T-t)+g(T-t)\bxi^{\rm G}(t)
\ee
where the central ingredient is the {\it score function} $\nabla\log\,p(\vec{x},t)$, which directs the $\vec{X}$-process towards $p_{\rm data}$, and contains the PDF $p$ of the forward process. Describing the backwards evolution in terms of a PDF $\vec{X}(t)\sim \tilde{p}(\vec{x},t)$ and starting from $\tilde{p}(\vec{x},0)=p(\vec{x},T)$, we have $\tilde{p}(\vec{x},t)=p(\vec{x},T-t)$, i.e., the score function ensures that $\vec{X}$ at any time $t$ is equal in distribution to $\vec{Y}$ at time $T-t$. The challenge of sampling the starting point $\vec{X}(0)\sim p(\vec{x},T)$ is simplified by choosing $\vec{f},g$ such that $p(\vec{x},t|\vec{x}')$ converges for $t\to \infty$ to a well-defined stationary distribution $p_{\rm st}(\vec{x})$, for example a simple Gaussian. For sufficiently large $T$, we have then $p(\vec{x},T)\approx p_{\rm st}(\vec{x})$ from which $\vec{X}(0)$ is sampled.

The generative process can be equivalently described by the deterministic probability flow ODE \cite{Song:2021aa}
\be
\label{pf}
\dot{\vec{X}}(t)=-\vec{f}(\vec{X},T-t)+\frac{D}{2}g^2(T-t)\nabla\log\,p(\vec{X},T-t),
\ee
which generates the same distribution $\tilde{p}(\vec{x},t)$ as Eq.~(\ref{reverse}) provided the same initial distribution  $\vec{X}(0)\sim p(\vec{x},T)$ is used.

Of course, if $p_{\rm data}$ is unknown, the exact analytical form of $p(\vec{x},t)$ and thus the score function is also not known. However, the generative power of this approach stems from the fact that the score function can be estimated from the available data (samples from $p_{\rm data}$) and thus yields an approximation of the exact sampling processes Eqs.~(\ref{reverse},\ref{pf}). One can show that the parameters $\btheta$ of a given parametric form $\vec{s}_{\btheta}$ for the score function parametrized for example by a neural network, can be estimated by optimizing the loss function
\be
\label{Gloss}
J_{\rm G}(\btheta)&=&\int_0^T\D t\,w(t)\mathbb{E}_{\vec{Y}_0\sim p_{\rm data}}\mathbb{E}_{\vec{Y}(t)|\vec{Y}_0}\left[\|\vec{s}_{\btheta}(\vec{Y},t)-\nabla\log\, p(\vec{Y},t|\vec{Y}_0\|^2\right]
\ee
with weight function $w(t)$. Eq.~(\ref{Gloss}) is called the {\em denoising score matching} (DSM) loss function \cite{Vincent:2011aa} and highlights that the estimation of $\vec{s}_{\btheta}$ does not require knowledge of the exact $p_{\rm data}$, but can be performed from finite samples of $p_{\rm data}$ using Monte-Carlo integration.

As discussed in \cite{Song:2021aa}, this SDE formulation unifies widely-used generative modelling approaches based on Markov chains, which correspond to discrete time approximations of Eqs.~(\ref{forward},\ref{reverse}). For example, discretizing $t,\vec{X}(t),g(T-t)$ over $[0,T]$ with time step $\Delta t$ and setting $\vec{f}=\vec{0}$, $D=1$ yields Eq.~(\ref{reverse}) as the iterative equation
\be
\label{smld}
\vec{X}_{i+1}=\vec{X}_i+g_i^2\vec{s}_{\btheta}(\vec{X}_i,T-t_i)\Delta t+g_i\sqrt{\Delta t}\,\vec{Z}_i,
\ee
where $\vec{Z}_i\sim\mathcal{N}(0,\vec{I})$ and $\vec{s}_{\btheta}(\vec{X}_i,T-t_i)$ approximates $\nabla\log\,p(\vec{X}_i,T-t_i)$. Eq.~(\ref{smld}) corresponds to SDM Langevin dynamics \cite{Song:2019aa} that generates samples of $p_{\rm data}$ by gradient descent plus noise over different noise scales $g_i$. Likewise, one can recover DDPM by setting $\vec{f}(\vec{y},t)=-\frac{1}{2}\beta(t)\vec{y}$, $g(t)=\sqrt{\beta(t)}$ and discretizing Eq.~(\ref{reverse}) \cite{Song:2021aa}.

\section{From diffusions to jump-diffusions}

The aim of this work is to consider the case when the noise $\bxi$ in Eq.~(\ref{forward}) is not given as purely Gaussian white noise, but contains non-Gaussian contributions in the form of discontinuous jumps that occur with a fixed rate and whose amplitudes are drawn independently from a given distribution. The general class of such processes with stationary and independent increments is known as L\'evy processes and have a well-developed mathematical theory \cite{Cont:2003aa}. In the following, I focus on the class of {\em finite activity L\'evy processes}, for which the jump amplitude distribution is normalizable, excluding the case of power-law or gamma distributed jump amplitudes. The mathematical theory is considerably simpler in this case and leads to a straightforward generalization of the conventional Gaussian case discussed in Sec.~\ref{Sec:intro}. The special case of pure power-law jumps corresponding to $\alpha$-stable noise with $1<\alpha\le 2$ has been discussed in detail in the LIM framework \cite{Yoon:2023aa} using rigorous methods.

Including jumps, the noise in Eq.~(\ref{forward}) is now expressed as
\be
\label{jnoise}
\bxi(t)=\bxi^{\rm G}(t)+\sum_{j=1}^{N_T}\vec{A}_j\delta(t-\tau_j),
\ee
where $\bxi^{\rm G}(t)$ is the Gaussian white noise as before and the jumps are described as follows: the number of jumps in the time interval $[0,T]$ follows a Poisson process with rate $\lambda$, i.e., $P(N_T=n)=\frac{(\lambda T)^n}{n!}e^{-\lambda T}$, the jump instances $\tau_j$ are uniform in $[0,T]$, and the jump amplitudes $A_j$ are drawn i.i.d. from a distribution $\rho$. Considering a finite time step $\Delta t$, one can show that the statistics of the noise increments $\Delta\bxi=\int_t^{t+\Delta t}\bxi(t')\D t'$ is then described by the characteristic function \cite{Cont:2003aa}:
\be
\label{xicf}
G_{\Delta\bxi}(\vec{k})=\mathbb{E}\left[e^{i\vec{k}{\cdot}\Delta\bxi}\right]=e^{-\psi(k)\Delta t},\qquad\qquad \psi(k)=\frac{D}{2}k^2-\lambda\phi(k)
\ee
with $k=\|\vec{k}\|$ and
\be
\label{phi}
\phi(k)=\int\D \vec{z}\left(e^{i\vec{k}{\cdot}\vec{z}}-1\right)\rho(\vec{z}).
\ee
contains the characteristic function of the jumps. A crucial constraint is that I assume isotropic jump distributions such that $\rho$ is a function of $z=\|\vec{z}\|$ only and, as a consequence, $\phi$ can be expressed as a function of $k=\|\vec{k}\|$ only. {Moreover, $\rho$ needs to be normalizable: $\int\D\vec{z}\,\rho(\vec{z})=1$.}

The following jump-diffusion generalization of Eqs.~(\ref{reverse},\ref{pf},\ref{Gloss}) is obtained when the noise of Eq.~(\ref{forward}) is specified by Eqs.~(\ref{xicf},\ref{phi}):
\begin{enumerate}
\item {\em Generalized score function}: Instead of $\nabla\log\,p(\vec{x},t)$, the score function that governs the generative process is now given by
\be
\label{gensc}
{\mathcal{S}(\vec{x},t)=\frac{\nabla\mathcal{V}(\vec{x},t)}{p(\vec{x},t)}}
\ee
where $p(\vec{x},t)$ is the PDF of the forward process $\vec{Y}(t)$ with Fourier transform (FT) $\hat{p}(\vec{k},t)$ and $\mathcal{V}$ is expressed in terms of the inverse FT
\be
\label{jdscore}
\mathcal{V}(\vec{x},t)&=&\mathcal{F}^{-1}\left\{\hat{p}(\vec{k},t)\frac{\psi(k\, g(t))}{k^2}\right\}
\ee
Eq.~(\ref{jdscore}) recovers the conventional score function for $\lambda=0$. Likewise, setting $\phi(k)=-\frac{a^2}{2}k^2$ in Eq.~(\ref{xicf}), i.e., additional Gaussian noise with intensity $D_2=\lambda a^2$ just recovers the conventional case with noise intensity $D\to D+D_2$.

\item {\em Generative process}: The jump-diffusion equivalent of the probability flow ODE Eq.~(\ref{pf}) is given by
\be
\label{jdpf}
{\dot{\vec{X}}(t)=-\vec{f}(\vec{X},T-t)+\mathcal{S}(\vec{X},T-t)}
\ee
The PDF of the generative process can likewise be generated by an SDE generalizing Eq.~(\ref{reverse})
\be
\label{jdsde}
{\dot{\vec{X}}(t)=-\vec{f}(\vec{X},T-t)+2\,\mathcal{S}(\vec{X},T-t)+g(T-t)\bxi(t)}
\ee
For the derivation of Eqs.~(\ref{jdscore},\ref{jdpf},\ref{jdsde}) see Appendix~\ref{Sec:reversal}

\item {\em Denoising score matching loss function}: The jump-diffusion equivalent of Eq.~(\ref{Gloss}) is given by
\be
\label{jdloss}
J_{\rm JD}(\btheta)&=&\int_0^T\D t\,w(t)\mathbb{E}_{\vec{Y}_0\sim p_{\rm data}}\mathbb{E}_{\vec{Y}(t)|\vec{Y}_0}\left[\left\|\vec{s}_{\btheta}(\vec{Y},t)-{\mathcal{S}(\vec{Y},t|\vec{Y}_0)}\right\|^2\right]
\ee
{Here, I introduce the conditional generalized score function $\mathcal{S}(\vec{x},t|\vec{x}')=\frac{\nabla\mathcal{V}(\vec{x},t|\vec{x}')}{p(\vec{x},t|\vec{x}')}$, where}
\be
\label{jdscorecond}
\mathcal{V}(\vec{x},t|\vec{x}')&=&\mathcal{F}^{-1}\left\{\hat{p}(\vec{k},t|\vec{x}')\frac{\psi(k\, g(t))}{k^2}\right\}
\ee
and $p(\vec{x},t|\vec{x}')$ denotes the conditional PDF of $\vec{Y}(t)$, Eq.~(\ref{forward}), with FT $\hat{p}(\vec{k},t|\vec{x}')$. Eq.~(\ref{jdloss}) can be derived from the explicit score matching loss function following standard steps, see Appendix~\ref{Sec:loss}

\item {\em Discrete approximations}: Eq.~(\ref{jdsde}) can be straightforwardly discretized leading to jump-diffusion generalizations of Markov chain diffusion models. For example, the jump-diffusion equivalent of Eq.~(\ref{smld}) with $\vec{f}=\vec{0}$ is given by
\be
\label{jdsm}
\vec{X}_{i+1}=\vec{X}_i+g_i^2\vec{s}_{\btheta}(\vec{X}_i,T-t_i)\Delta t+g_i\sqrt{\Delta t}\,\vec{Z}_i+g_i\Delta\vec{J}_i
\ee
where $\vec{s}_{\btheta}$ is estimated by Eq.~(\ref{jdloss}) and $\vec{Z}_i\sim\mathcal{N}(\vec{0},D\vec{I})$. The jump increment $\Delta \vec{J}_i$ is generated by drawing $M\sim \text{Poisson}(\lambda\Delta t)$ and setting $\Delta\vec{J}_i=\sum_{j=1}^M\vec{A}_j$, where $\vec{A}_j\sim \rho$ i.i.d. One obtains likewise a jump-diffusion variant of DDPM by setting $\vec{f}(\vec{x},t)=-\frac{1}{2}\beta(t)\vec{x}$, $g(t)=\sqrt{\beta(t)}$.

\end{enumerate}

One of the key insights here is that changing the noise type in the forward process leads to a modification of the score function for consistent denoising according to the correct time-reversal. In particular the DSM formulation Eq.~(\ref{jdsm}) is conceptually different from an heuristic approach that simply samples Langevin dynamics with the standard score function $\nabla\log\,p(\vec{x},t)$ plus non-Gaussian noise. I now discuss the detailed implementation of the framework for a specific choice of $\vec{f},g$, and jump amplitude distribution.

\subsection{The JL model: Laplace distributed jump amplitudes}

{For efficient generative modelling, it is desirable to have: (i) a closed analytical form of the conditional score function, which facilitates the loss function minimization; and (ii) to be able to generate samples from $p_{\rm st}(\vec{x})$ efficiently. For standard DMs, a common choice is thus an Ornstein-Uhlenbeck process as forward process, where $p(\vec{x},t|\vec{x}')$ is a Gaussian and thus satisfies both requirements. In the jump-diffusion case, I consider likewise $\vec{Y}(t)$ as a (generalized) Ornstein-Uhlenbeck process, where for simplicity
\be
\label{OUmodel}
\vec{f}(\vec{x},t)=-\frac{1}{2}\,\vec{x}\qquad,\qquad\qquad g(t)=1
\ee
The choice of amplitude distribution of course significantly affects the conditional PDF and generalized score function. The non-Gaussian model becomes particularly simple for the case of a pure jump process ($D=0$) with $\lambda=1$ and amplitudes drawn from an isotropic multivariate Laplace distribution: $\vec{A}_j\sim \text{L}_d(\sigma^2)$. The distribution $\text{L}_d(\sigma^2)$ not only represents a multivariate extension of the exponential distribution and is thus a standard case to consider beyond the Gaussian, but also allows for straightforward sampling. The resulting diffusion model is given in closed analytical form and is only slightly more computationally costly than the standard Gaussian case. I refer to this model as {\em jump Laplace (JL) model}.}

{For details of the derivation I refer to Appendix~\ref{Sec:genOU}. The conditional PDF of the JL model is given by
\be
\label{OUprop}
p(\vec{x},t|\vec{x}')&=&\rho_{\rm L}(\vec{x}-\vec{x}'e^{-t/2},\sigma)\left(1-e^{-t}\right)+\delta(\vec{x}-\vec{x}'e^{-t/2})e^{-t}
\ee
where
\be
\label{G1jl}
\rho_{\rm L}(\vec{x},\sigma)=\frac{2}{(\sigma)^d(2\pi)^{d/2}}\left(\frac{\|\vec{x}\|}{\sqrt{2}\,\sigma}\right)^{1-\frac{d}{2}}K_{1-\frac{d}{2}}\left(\frac{\|\vec{x}\|\sqrt{2}}{\sigma}\right)
\ee
is the distribution $\text{L}_d(\sigma^2)$ itself (see Appendix~\ref{Sec:laplace}). The conditional generalized score function is
\be
\label{jlscore}
\mathcal{S}(\vec{x},t|\vec{x}')=\frac{\vec{x}-\vec{x}'e^{-t/2}}{\|\vec{x}-\vec{x}'e^{-t/2}\|}\mathcal{G}(\|\vec{x}-\vec{x}'e^{-t/2}\|,t)
\ee
introducing the function
\be
\label{G2jl}
\mathcal{G}(x,t)=-\frac{1}{2}x-\frac{\sigma}{\sqrt{2}}\frac{K_{d/2}\left(\frac{x\sqrt{2}}{\sigma}\right)}{K_{1-d/2}\left(\frac{x\sqrt{2}}{\sigma}\right)}\left(\frac{e^{-t}}{1-e^{-t}}\right),\qquad\qquad x>0
\ee
with $\mathcal{G}(x,t)=0$ when $x=0$. In Eqs.~(\ref{G1jl},\ref{G2jl}), $K_\nu$ denotes the modified Bessel function of the 2nd kind. Moreover, the stationary distribution is just
\be
p_{\rm st}(\vec{x})=\lim_{t\to\infty}p(\vec{x},t|\vec{x}')=\rho_{\rm L}(\vec{x},\sigma)
\ee
and can be sampled like the jumps.
}

\subsection{Implementation of the JL model}

{The implementation of the JL model thus requires the following steps:

1. {\em Estimating the generalized score function.} Optimizing the loss function Eq.~(\ref{jdloss}) with $\mathcal{S}(\vec{x},t|\vec{x}')$ given in Eq.~(\ref{jlscore}) yields $\vec{s}_{\btheta}$ as approximation of the exact $\mathcal{S}(\vec{x},t)$. As in the Gaussian case, the numerical stability of the training process can be improved by scaling the time-dependence in the score function. Eq.~(\ref{G2jl}) indicates that the divergence in $\mathcal{S}(\vec{x},t|\vec{x}')$ for $t\to 0$ can be removed by multiplying with $1-e^{-t}$, thus the loss function becomes
\begin{equation}
\label{jdloss_scl}
J_{\rm JD}(\btheta)=\int_0^T\D t\frac{w(t)}{(1-e^{-t})^2}\mathbb{E}_{\vec{Y}_0\sim p_{\rm data}}\mathbb{E}_{\vec{Y}(t)|\vec{Y}(0)}\left[\left\|\hat{\vec{s}}_{\btheta}(\vec{Y},t)-(1-e^{-t})\mathcal{S}(\vec{Y},t|\vec{Y}_0)\right\|^2\right]
\end{equation}
Direct integration of Eqs.~(\ref{forward},\ref{jnoise}) with $\vec{f}(\vec{x},t)=-\frac{1}{2}\vec{x}$, $g(t)=1$, $D=0$ yields
\be
\label{Yincr}
\vec{Y}(t)=\vec{Y}(0)e^{-t/2}+\vec{J}(t)
\ee
where the jump increment is derived as
\be
\label{incrdef}
\vec{J}(t)=\sum_{j=1}^{N_t}\vec{A}_je^{-(t-\tau_j)/2}
\ee
Eqs.~(\ref{Yincr},\ref{incrdef}) allow for exact sampling of $\vec{Y}(t)$ without discretization errors, see Section~\ref{Sec:jl_sampling} below for details. Setting the weight function $w(t)=(1-e^{-t})^2/T$ in Eq.~(\ref{jdloss_scl}) then yields the loss function
\begin{equation}
\label{jdloss_scl2}
J_{\rm JD}(\btheta)=\mathbb{E}_{t\sim{\rm U}(0,T)}\mathbb{E}_{\vec{Y}_0\sim p_{\rm data}}\mathbb{E}_{\vec{J}(t)}\left[\left\|\hat{\vec{s}}_{\btheta}(\vec{Y}_0e^{-t/2}+\vec{J}(t),t)-\frac{\vec{J}(t)}{\|\vec{J}(t)\|}\hat{\mathcal{G}}(\|\vec{J}(t)\|,t)\right\|^2\right]
\end{equation}
using Eqs.~(\ref{jlscore},\ref{Yincr}) and defining $\hat{\mathcal{G}}(x,t)=(1-e^{-t})\mathcal{G}(x,t)$.

2. {\em Sampling of the generative process.} Implementing the generative process upon replacing $\mathcal{S}$ by $\vec{s}_{\btheta}$ via the probability flow ODE Eq.~(\ref{jdpf}) or SDE Eq.~(\ref{jdsde}) can be made more efficient with exponential samplers \cite{Zhang:2023aa} that exploit the linearity akin to Eq.~(\ref{Yincr}). For Eq.~(\ref{jdpf}), direct integration of the linear term while keeping $\vec{s}_{\btheta}$ constant yields the simple discretization:
\be
\label{ODEdiscr}
\vec{X}_{i+1}&=&\vec{X}_ie^{\Delta t/2}+2\left(e^{\Delta t/2}-1\right)\vec{s}_{\btheta}(\vec{X}_i,T-t_i)\nonumber\\
&=&\vec{X}_ie^{\Delta t/2}+2\left(\frac{e^{\Delta t/2}-1}{1-e^{-(T-t_i)}}\right)\hat{\vec{s}}_{\btheta}(\vec{X}_i,T-t_i)
\ee
using the rescaled estimated score function introduced in Eq.~(\ref{jdloss_scl}). For Eq.~(\ref{jdsde}) direct integration yields likewise
\be
\label{SDEdiscr}
\vec{X}_{i+1}=\vec{X}_ie^{\Delta t/2}+4\left(\frac{e^{\Delta t/2}-1}{1-e^{-(T-t_i)}}\right)\hat{\vec{s}}_{\btheta}(\vec{X}_i,T-t_i))+\tilde{\vec{J}}(\Delta t),
\ee
where the (backwards) jump increment $\tilde{\vec{J}}(t)$ is derived as 
\be
\label{incrdef2}
\tilde{\vec{J}}(t)=\sum_{j=1}^{N_t}\vec{A}_je^{(t-\tau_j)/2}
\ee
In both Eqs.~(\ref{ODEdiscr},\ref{SDEdiscr}), the starting point is sampled $\vec{X}_0\sim \text{L}_d(\sigma^2)$ since $p_{\rm st}(\vec{x})=\rho_{\rm L}(\vec{x},\sigma)$.

 \subsection{Sampling algorithms}
 \label{Sec:jl_sampling}

The JL model requires the sampling of three different random variates:

\begin{enumerate}
\item[(a)] $\vec{X}_0\sim \text{L}_d(\sigma^2)$. As outlined in Appendix~\ref{Sec:laplace}, random variates of $\text{L}_d(\sigma^2)$ are generated by setting
\be
\label{pstsample}
\vec{X}_0=\sqrt{\Gamma}\,\vec{Z}
\ee
where $\Gamma\sim \text{Gamma}(1)$ and $\vec{Z}\sim \mathcal{N}(\vec{0},\sigma^2\vec{I})$.

\item[(b)] Jump increment $\vec{J}(t)$. The definition Eq.~(\ref{incrdef}) can be implemented directly by: (i) Drawing $M\sim \text{Poisson}(\lambda t)$, where $\lambda=1$. (ii) Determining $M$ times $\tau_j\sim \text{U}(0,t)$ and $\vec{A}_j\sim \text{L}_d(\sigma^2)$. (iii) Setting $\vec{J}(t)=\sum_{j=1}^M\vec{A}_je^{-(t-\tau_j)/2}$. However, a simpler method is found by noting that the distribution of $\vec{J}(t)$ is simply (compare Eqs.~(\ref{OUprop},\ref{Yincr}))
\be
\label{Jdist}
p(\vec{J},t)=\rho_{\rm L}(\vec{J},\sigma)\left(1-e^{-t}\right)+\delta(\vec{J})e^{-t}
\ee
Thus we can sample $\vec{J}(t)$ by selecting $\omega\sim U(0,1)$ and setting $\vec{J}\sim \text{L}_d(\sigma^2)$ if $\omega>e^{-t}$ and $\vec{J}=\vec{0}$ otherwise.

\item[(c)] Jump increment $\tilde{\vec{J}}(t)$. The definition Eq.~(\ref{incrdef2}) indicates that the same algorithm as in (b) applies upon changing (iii) to $\tilde{\vec{J}}(t)=\sum_{j=1}^M\vec{A}_je^{(t-\tau_j)/2}$. Interestingly, a simplified algorithm can here also be found. As derived in Appendix~\ref{Sec:jtilde}, the PDF of $\tilde{\vec{J}}(t)$ is given by
\be
p(\tilde{\vec{J}},t)=\rho_{\rm L}(\tilde{\vec{J}},\sigma\,e^{t/2})\left(1-e^{-t}\right)+\delta(\tilde{\vec{J}})e^{-t}
\ee
The random variates $\tilde{\vec{J}}(t)$ can thus be sampled by selecting $\omega\sim U(0,1)$ and setting $\tilde{\vec{J}}\sim \text{L}_d(\sigma^2e^t)$ if $\omega>e^{-t}$ and $\tilde{\vec{J}}=\vec{0}$ otherwise.

\end{enumerate}
}

\section{Connection with other works}

{\bf LIM.} It is instructive to compare the results in this work with those of LIM \cite{Yoon:2023aa}, where the forward process Eq.~(\ref{forward}) is driven by $\alpha$-stable L\'evy noise with $1<\alpha\le 2$. In this case, a modification of the standard score function has also been obtained leading to the fractional score function \cite{Yoon:2023aa}
\be
\label{Salpha}
\mathcal{S}_\alpha(\vec{x},t)=\frac{\Delta^{\frac{\alpha-2}{2}}\nabla p(\vec{x},t)}{p(\vec{x},t)}
\ee
where $\Delta^{\frac{\alpha-2}{2}}$ denotes the fractional Laplacian of order $\frac{\alpha-2}{2}$. While the derivation of the jump-diffusion generative process in Appendix~\ref{Sec:reversal} is not valid for this case, one can naively consider $\alpha$-stable noise in the generalized score function by substituting the corresponding $\psi(k)$ in Eq.~(\ref{jdscore}). Indeed, the characteristic function of isotropic $\alpha$-stable noise increments $\Delta\bxi$ with $1<\alpha\le 2$ is likewise given by Eq.~(\ref{xicf}), but with $\psi(k)=k^\alpha$ \cite{Cont:2003aa}. For this special case, one finds from Eq.~(\ref{jdscore})
\be
\nabla\mathcal{V}(\vec{x},t)&=&g(t)^\alpha\nabla\mathcal{F}^{-1}\left\{\hat{p}(\vec{k},t)k^{\alpha-2}\right\} \nonumber\\
&=&g(t)^\alpha\Delta^{\frac{\alpha-2}{2}}\nabla p(\vec{x},t)\label{gradvalpha}
\ee
and the generalized score function $\mathcal{S}(\vec{x},t)$ of Eq.~(\ref{gensc}) recovers the fractional score function Eq.~(\ref{Salpha}). The generative process Eq.~(\ref{jdpf}) then yields a probability flow ODE for LIM. These results are somewhat surprising, since the rigorous analysis of the time-reversal of Markov jump processes in \cite{Conforti:2022aa} shows that in the case of $\alpha$-stable L\'evy processes with $1<\alpha\le2$, a very different time-reversal formula applies than for finite activity L\'evy processes. However, the result in Eq.~(\ref{gradvalpha}) suggests that the generalized score function Eq.~(\ref{gensc}) might be valid more generally.

{\bf Generator matching.} The generator matching framework of \cite{Holderrieth:2025aa} specifies the backward process in terms of the generator of a general Markov process and learns its coefficients using a prescribed conditional probability distribution that maps between the simple distribution and $p_{\rm data}$ (called probability path). Like in flow-matching, the probability path replaces the forward noising process and is explicitly specified at model construction. The key insight in generator matching is that the probability path and conditional generator are related through the Kolmogorov-Forward equation. Thus the coefficients in the conditional generator can be determined once a path is given, although this is strictly only possible in $d=1$, while for $d>1$ a composition rule needs to be applied \cite{Holderrieth:2025aa}. 
\begin{enumerate}
\item The framework here is consistent with generator matching if we take as probability path the time reverse of the conditional PDF $p(\vec{x},t|\vec{x}')$ of the forward process Eqs.~(\ref{forward},\ref{jnoise}) and consider a backward generator containing a flow component only. The conditional flow field $\vec{u}(\vec{x},t|\vec{x}')$ then satisfies the continuity equation (i.e., the adjoint Kolmogorov-Forward equation in this case) with $\tilde{p}(\vec{x},t|\vec{x}')=p(\vec{x},T-t|\vec{x}')$:
\be
\frac{\partial}{\partial t}\tilde{p}(\vec{x},t|\vec{x}')=-\nabla\left[\vec{u}(\vec{x},t|\vec{x}')\tilde{p}(\vec{x},t|\vec{x}')\right]
\ee
Since $\tilde{p}(\vec{x},t|\vec{x}')$ satisfies the same Fokker-Planck equation as the marginal PDF, the derivation in Appendix~\ref{Sec:reversal} can be performed in the same way and yields
\be
\label{gmfield}
\vec{u}(\vec{x},t|\vec{x}')=-\vec{f}(\vec{x})+\mathcal{S}(\vec{x},T-t|\vec{x}')
\ee
where the conditional generalized score function is defined via Eq.~(\ref{jdscorecond}). Eq.~(\ref{gmfield}) recovers the ODE representation Eq.~(\ref{jlscore}). We see that generator matching reverses the procedure used here and yields the same results in arbitrary $d$ because the probability path is the solution of a given Fokker-Planck equation and thus the analytical results derived here can be used.

\item The SDE representation then likewise follows due to the fact that we can define a generator $\mathcal{L}_t^{\rm NG}$ satisfying
\be
\label{gmdiv}
\mathcal{L}_t^{\rm NG}\tilde{p}(\vec{x},t)&=&-\nabla\cdot\left[\mathcal{S}(\vec{x},T-t)\tilde{p}(\vec{x},t)\right]+\frac{D}{2}g^2(t)\nabla^2\tilde{p}(\vec{x},t)\nonumber\\
&&+\lambda\int\D\vec{z}\,\rho(z)\left[\tilde{p}(\vec{x}+g(t)\vec{z},t)-\tilde{p}(\vec{x},t)\right]\\
&=&0\label{zerogen}
\ee
where $\mathcal{S}(\vec{x},t)$ is defined in Eq.~(\ref{gensc}) and $\tilde{p}(\vec{x},t)=p(\vec{x},T-t)$. Eq.~(\ref{zerogen}) likewise follows from the results in Appendix~\ref{Sec:reversal}. In generator matching, a generator satisfying $\mathcal{L}_t\tilde{p}(\vec{x},t)=0$ is called a {\em divergence-less generator}, which can be added to a pure flow ODE model to turn it into a model with stochastic sampling as in the SDE representation Eq.~(\ref{jdsde}). Crucially, since the flow component is the same as for the ODE, this does not require estimation of additional coefficients. The generator Eq.~(\ref{gmdiv}) can be combined with other generators according to the generator matching composition rules to formulate more complex models.

\item It should be highlighted that, without a divergence-less generator, generator matching with jumps always contains an estimated jump distribution for the backward process, since the jumps need to be adapted to the chosen probability path. As shown in \cite{Holderrieth:2025aa}, even simple probability paths then lead to complicated jump distributions with space- and/or time-inhomogeneities.

\end{enumerate}

\section{Numerical experiments}

\begin{figure}
\includegraphics[width=\textwidth]{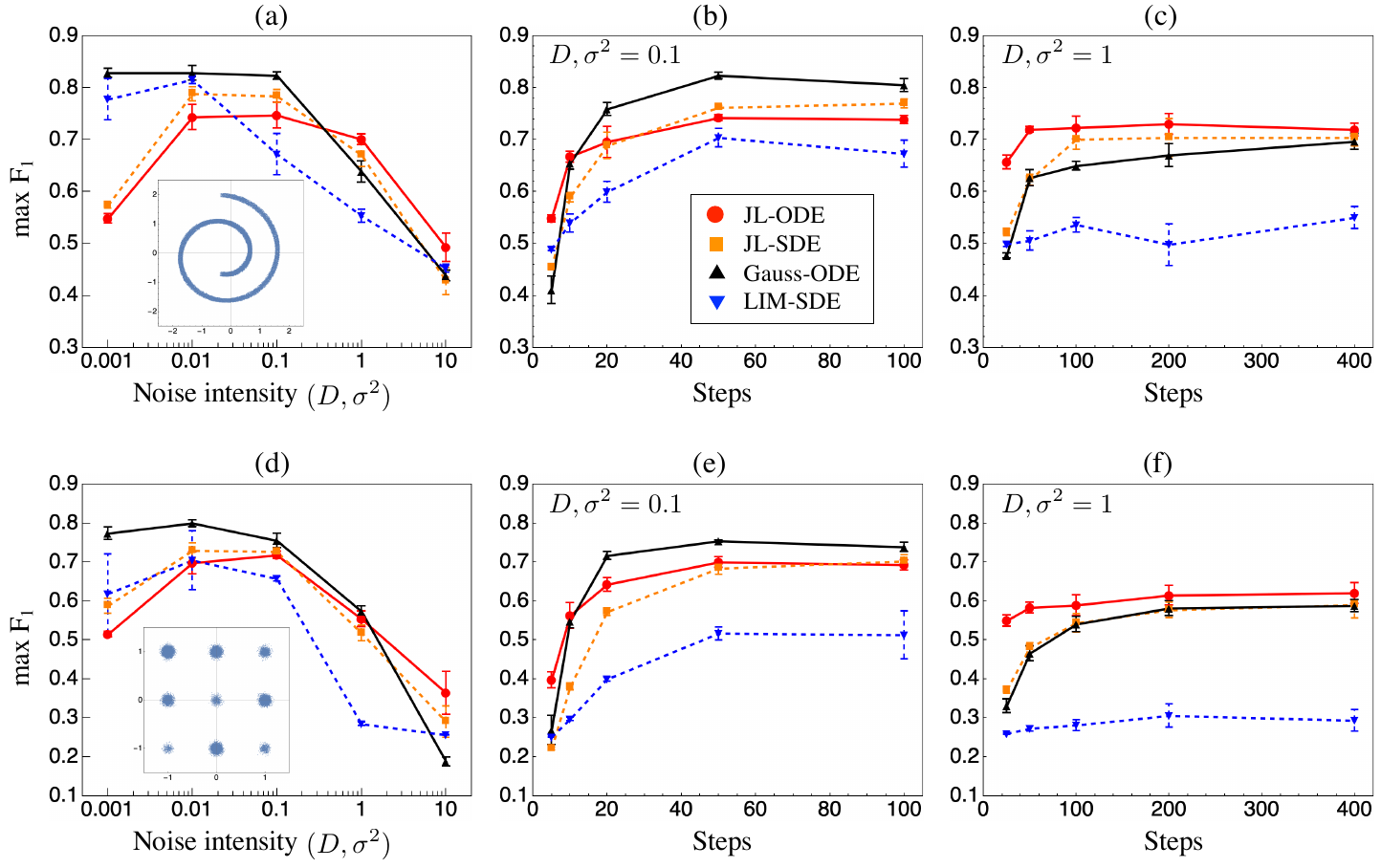}
\caption{\label{Fig:test}{Numerical experiments in $d=2$: (a,b,c) $p_{\rm data}$ given as a standardized swiss roll distribution; (d,e,f) $p_{\rm data}$ given as a Gaussian mixture. Plotted is the $F_1$ metric that measures the accuracy of the generated distributions. While the best performance is achieved by the Gauss-ODE method with $\max F_1>80$ in both tests, the JL model performs better at high noise intensities and for small step numbers. JL performs consistently better than LIM-SDE apart from the regime of very low noise intensities. Protocol details: $\alpha=1.9$ for LIM-SDE, $10^5$ training samples, $2\cdot 10^4$ test samples, $T=10$, error bars given as the standard error from 5 repetitions, sampling steps $=100$ in (a,d). The neural network is a simple multilayer perceptron with 4 layers, 128 nodes, and GELU activations. For more details on the implementation and the Gauss-ODE and LIM-SDE methods, see Appendix~\ref{Sec:exp}.}}
\end{figure}

{In order to demonstrate the applicability of the framework, I consider numerical experiments on two simple data distributions in $d=2$: (a) a standardized swiss roll distribution; and (b) a mixture of 9 Gaussians with varying weights arranged on a grid. The focus is here to understand the effect of non-Gaussian jumps on the accuracy of sample generation compared with an equivalent Gaussian DM. Crucially, the underlying forward process, training protocol, and neural network architecture are intentionally kept very simple to disentangle effects due to specific implementations and those due to the noise type only. Indeed, it is likely that non-Gaussian DMs require different types of protocols/architectures to perform optimally compared with standard DMs, whose designs have been optimized for various applications in recent years. 

Fig.~\ref{Fig:test} shows results for the JL model (both ODE and SDE formulations), the corresponding Gaussian DM as a probability flow ODE, and an equivalent LIM SDE of \cite{Yoon:2023aa}, which is driven by isotropic $\alpha$-stable noise. More details on the numerical implementations are presented in Appendix~\ref{Sec:exp}. As performance metric I choose the $F_1$ metric, which is based on an extension of the notions of precision and recall to continuous distributions \cite{Sajjadi:2018aa}, and Fig.~\ref{Fig:test} shows the maximal $F_1$ values for samples generated by the four different methods. 

In Figs.~\ref{Fig:test}(a,d) results for fixed number of sampling steps ($=100$) and different noise intensities $D,\sigma^2=0.001-10$ are shown. For LIM-SDE the noise intensity is undefined, but scaling the noise term by $(\sigma^2/2)^{1/\alpha}$ ensures that for $\alpha=2$, i.e., when the $\alpha$-stable noise converges to Gaussian white noise, the resulting noise intensity is $\sigma^2$. The non-Gaussian methods all exhibit non-monotonic behaviour, where optimal performance is achieved in the range $\sigma^2=0.01- 0.1$, but still below that of the Gaussian model at same noise intensity. For larger noise intensity the accuracy of all methods drops considerably. Interestingly, we can identify a parameter regime, where JL-ODE performs significantly better than Gauss-ODE, which is at $D,\sigma^2=1$ (Fig.~\ref{Fig:test}(c,f)) and small number of sampling steps. For steps$=25$, JL-ODE achieves $\max F_1$ of 0.66 for the swiss roll and 0.55 for the Gaussian mixture, while $\max F_1<0.52$ and $<0.37$, respectively, for the other methods. For $D,\sigma^2<1$ on the other hand, Gauss-ODE performs best throughout, achieving the highest absolute values of $\max F_1\ge 0.8$ in both tests. Comparing the JL models and LIM-SDE, we see that LIM-SDE performs better at very low noise intensities and also achieves higher absolute $\max F_1$ values, but is consistently outperformed at larger noise intensities. In particular, for $D,\sigma^2=1$, LIM-SDE performs quite poorly compared with the JL models. It should be noted that the same Gaussian mixture test of Fig.~\ref{Fig:test}(d,e,f) was first implemented in \cite{Shariatian:2025aa}, where LIM-SDE was also tested and achieved $\max\,F_1=0.88$ for $\alpha=1.9$, but for a more sophisticated neural network architecture. The DLPM method of \cite{Shariatian:2025aa} achieved $\max\,F_1=0.94$, but its performance under the test conditions here was not evaluated.

}

\section{Conclusion}

Extending the noise in the forward process to jump-diffusions leads to a straightforward non-Gaussian generalization of the standard DM framework of \cite{Song:2021aa}, {which is likewise based on a MSE loss}. The results highlight that the score function needs to be adapted to the noise and only takes the conventional form for the Gaussian noise case, while it is otherwise determined by the non-Gaussian details of the noise. {The implementation of the JL model rivals both the LIM and standard Gaussian model in its analytical and computational simplicity and might lead to performance improvements in specific applications. In particular the fact that it performs quite well for a small number of discretization steps as shown in Fig.~\ref{Fig:test}(c,f) seems promising.} Apart from investigating the performance of the framework in more realistic scenarios such as image synthesis, further detailed analysis is also needed to understand the effects of the various parameters and protocols underlying the implementation. {It should be highlighted that the design space of optimal protocols and network architectures is likely different from that of Gaussian models, so a lot more work is needed to realize the potential impact of non-Gaussian models.}

{On the theoretical side, in addition to extending the framework to more general Markov processes conceptually, it would be interesting to investigate other specific jump amplitude distributions and the combined effect of Gaussian white noise and jumps in detailed implementations. While the generalized Ornstein-Uhlenbeck process is generally integrable in Fourier space (see Appendix~\ref{Sec:genOU}) the resulting expressions for the conditional PDF and generalized score function are typically not available in closed form. It might thus be worthwhile to consider other score matching approaches such as, e.g., sliced score matching \cite{Song:2020aa} and investigate how these can be adapted to the non-Gaussian case. }


\newpage

\appendix

\section{Time-reversal of jump-diffusion processes}
\label{Sec:reversal}

Before we look in more detail into the jump-diffusion generalization of the standard generative framework with Gaussian noise outlined in Sec.~\ref{Sec:intro}, a heuristic derivation of these results is presented. The aim is to derive an equation for the PDF $\tilde{p}(\vec{x},t)$ of the time-reverse or generative process. This PDF can be specified by the condition that $\tilde{p}(\vec{x},t)=p(\vec{x},T-t)$ for $t\in [0,T]$. Using the Fokker-Planck equation associated with Eq.~(\ref{forward}) we obtain then
\be
\frac{\partial}{\partial t}\tilde{p}(\vec{x},t)&=&\frac{\partial}{\partial t}p(\vec{x},T-t)\nonumber\\
&=&-\left(-\nabla{\cdot}\vec{f}(\vec{X},T-t)p(\vec{x},T-t)+\frac{D}{2} g^2(T-t)\nabla^2p(\vec{x},T-t)\right)\nonumber\\
&=&\nabla{\cdot}\vec{f}(\vec{X},T-t)\tilde{p}(\vec{x},t)-\frac{D}{2} g^2(T-t)\nabla^2\tilde{p}(\vec{x},t)\label{deriv1}
\ee
Eq.~(\ref{deriv1}) can be expressed further in the form
\be
\frac{\partial}{\partial t}\tilde{p}(\vec{x},t)&=&\nabla{\cdot}\vec{f}(\vec{X},T-t)\tilde{p}(\vec{x},t)-D g^2(T-t)\nabla^2\tilde{p}(\vec{x},t)+\frac{D}{2} g^2(T-t)\nabla^2\tilde{p}(\vec{x},t)\nonumber\\
&=&\nabla{\cdot}\left(\vec{f}(\vec{X},T-t)-D g^2(T-t)\nabla\log\,\tilde{p}(\vec{x},t)\right)\tilde{p}(\vec{x},t)+\frac{D}{2} g^2(T-t)\nabla^2\tilde{p}(\vec{x},t)\nonumber\\
&=&\nabla {\cdot}\left(\vec{f}(\vec{x},T-t)-Dg^2(T-t)\nabla\log\,p(\vec{x},T-t)\right)\tilde{p}(\vec{x},t)\nonumber\\
&&+\frac{D}{2} g^2(T-t)\nabla^2\tilde{p}(\vec{x},t)\label{revfpe}
\ee
Eq.~(\ref{revfpe}) is the Fokker-Planck equation corresponding to the SDE Eq.~(\ref{reverse}), which thus represents a stochastic process with PDF $\tilde{p}(\vec{x},t)$. The initial condition needs to satisfy $\vec{X}(0)\sim \tilde{p}(\vec{x},0)=p(\vec{x},T)$.

Rewriting the last term of Eq.~(\ref{revfpe}) with $\tilde{p}(\vec{x},t)=p(\vec{x},T-t)$ yields further 
\be
\frac{\partial}{\partial t}\tilde{p}(\vec{x},t)&=&\nabla {\cdot}\left(\vec{f}(\vec{x},T-t)-Dg^2(T-t)\nabla\log\,p(\vec{x},T-t)\right.\nonumber\\
&&\left.+\frac{D}{2}g^2(T-t)\nabla\log\,p(\vec{x},T-t)\right)\tilde{p}(\vec{x},t)\nonumber\\
&=&-\nabla {\cdot}\left(-\vec{f}(\vec{x},T-t)+\frac{D}{2}g^2(T-t)\nabla\log\,p(\vec{x},T-t)\right)\tilde{p}(\vec{x},t)\label{liou}
\ee
Eq.~(\ref{liou}) is the continuity equation corresponding to the deterministic dynamics Eq.~(\ref{pf}), which likewise generates the PDF $\tilde{p}(\vec{x},t)$ for the initial condition $\vec{X}(0)\sim p(\vec{x},T)$. So we see that the generative process in the SDE description results from expressing the negative diffusion as a drift plus positive diffusion, while the ODE description results from expressing the negative diffusion as a drift only. We will see that the same heuristic prescription can also be applied for jump diffusions and agrees with rigorous results on the time-reversal of Markov processes with jumps, see Appendix~\ref{Sec:alt}.

The forward process is now specified by Eq.~(\ref{forward}) with noise $\bxi(t)$ characterized in Eqs.~(\ref{jnoise},\ref{xicf},\ref{phi}). The Fokker-Planck equation associated with $\vec{Y}(t)$ is obtained by standard methods \cite{Cont:2003aa} and has the integral form
\be
\label{jfpe}
\frac{\partial}{\partial t}p(\vec{x},t)&=&-\nabla {\cdot}\vec{f}(\vec{x},t)p(\vec{x},t)+\frac{D}{2}g^2(t)\nabla^2p(\vec{x},t)\nonumber\\
&&+\lambda\int\D\vec{z}\,\rho(z)\left[p(\vec{x}+g(t)\vec{z},t)-p(\vec{x},t)\right]
\ee
The probability flow formalism for L\'evy processes derived from Eq.~(\ref{jfpe}) has been investigated in \cite{Huang:2025aa}. I likewise use the identity
\be
\label{identity}
p(\vec{x}+g(t)\vec{z},t)-p(\vec{x},t)&=&\int_0^1\D \gamma\, \frac{\D}{\D\gamma} p(\vec{x}+\gamma g(t)\vec{z},t)\nonumber\\
&=&g(t)\int_0^1\D \gamma\, \vec{z}{\cdot}\nabla p(\vec{x}+\gamma g(t)\vec{z},t)
\ee 
such that Eq.~(\ref{jfpe}) can be rewritten as
\be
\frac{\partial}{\partial t}p(\vec{x},t)&=&-\nabla {\cdot}\vec{f}(\vec{x},t)p(\vec{x},t)+\frac{D}{2}g^2(t)\nabla^2p(\vec{x},t)\nonumber\\
&&+\lambda \,g(t)\int_0^1\D \gamma\int\D\vec{z}\,\rho(z)\vec{z}{\cdot} \nabla p(\vec{x}+\gamma g(t)\vec{z},t)\nonumber\\
&=&-\nabla{\cdot}\left(\vec{f}(\vec{x},t)-\frac{D}{2}g^2(t)\nabla \log\,p(\vec{x},t)\right.\nonumber\\
&&\left.-\lambda\, g(t)\int_0^1\D \gamma\int\D\vec{z}\,\vec{z}\,\rho(z) \frac{p(\vec{x}+\gamma g(t)\vec{z},t)}{p(\vec{x},t)}\right)p(\vec{x},t)\label{jfpe2}
\ee
Starting again from the condition $\tilde{p}(\vec{x},t)=p(\vec{x},T-t)$, an evolution equation for $\tilde{p}$ is derived as
\be
\frac{\partial}{\partial t}\tilde{p}(\vec{x},t)&=&\frac{\partial}{\partial t}p(\vec{x},T-t)\nonumber\\
&=&\nabla{\cdot}\left(\vec{f}(\vec{x},T-t)-\frac{D}{2}g^2(T-t)\nabla \log\,p(\vec{x},T-t)\right.\nonumber\\
&&\left.-\lambda\,g(T-t)\int_0^1\D \gamma\int\D\vec{z}\,\vec{z}\,\rho(z) \frac{p(\vec{x}+\gamma g(T-t)\vec{z},T-t)}{p(\vec{x},T-t)}\right)\tilde{p}(\vec{x},t)\nonumber\\
&=&-\nabla{\cdot}\left(-\vec{f}(\vec{x},T-t)+\frac{D}{2}g^2(T-t)\nabla \log\,p(\vec{x},T-t)\right.\nonumber\\
&&\left.+\lambda\frac{\blam(\vec{x},T-t)}{p(\vec{x},T-t)}\right)\tilde{p}(\vec{x},t)\label{genjfpe}
\ee
Here, I define the vector field
\be
\label{lambda}
\blam(\vec{x},t)=g(t)\int_0^1\D \gamma\int\D\vec{z}\,\vec{z}\,\rho(z) p(\vec{x}+\gamma g(t)\vec{z},t)
\ee
Below in Eq.~(\ref{Sec:gensc}), I will derive an explicit expression for $\blam$. Eq.~(\ref{genjfpe}) is the continuity equation for the generative process, which takes the form of the probability flow ODE
\be
\label{jpf}
\dot{\vec{X}}(t)=-\vec{f}(\vec{X},T-t)+\frac{D}{2}g^2(T-t)\nabla \log\,p(\vec{X},T-t)+\lambda\frac{\blam(\vec{X},T-t)}{p(\vec{X},T-t)}
\ee
The same result is obtained from rigorous results on the generator for time-reversed Markovian jump processes confirming the validity of Eq.~(\ref{jpf}), see Appendix~\ref{Sec:alt}.

In order to keep a jump operator in the evolution equation for $\tilde{p}$ I simply introduce a factor 2 akin to the procedure in the Gaussian case. Eq.~(\ref{genjfpe}) can be accordingly rewritten as
\be
\frac{\partial}{\partial t}\tilde{p}(\vec{x},t)&=&-\nabla{\cdot}\left(-\vec{f}(\vec{x},T-t)+Dg^2(T-t)\nabla \log\,p(\vec{x},T-t)+2\lambda\frac{\blam(\vec{x},T-t)}{p(\vec{x},T-t)}\right)\tilde{p}(\vec{x},t)\nonumber\\
&&+\frac{D}{2}g^2(T-t)\nabla^2\log\,p(\vec{x},T-t)\tilde{p}(\vec{x},t)+\lambda\nabla\blam(\vec{x},T-t)\nonumber\\
&=&-\nabla{\cdot}\left(-\vec{f}(\vec{x},T-t)+Dg^2(T-t)\nabla \log\,p(\vec{x},T-t)+2\lambda\frac{\blam(\vec{x},T-t)}{p(\vec{x},T-t)}\right)\tilde{p}(\vec{x},t)\nonumber\\
&&+\frac{D}{2}g^2(T-t)\nabla^2\tilde{p}(\vec{x},t)+\lambda\,g(T-t)\int_0^1\D \gamma\int\D\vec{z}\,\vec{z}\,\rho(z) \nabla \tilde{p}(\vec{x}+\gamma g(T-t)\vec{z},t)\nonumber\\
&=&-\nabla{\cdot}\left(-\vec{f}(\vec{x},T-t)+Dg^2(T-t)\nabla \log\,p(\vec{x},T-t)+2\lambda\frac{\blam(\vec{x},T-t)}{p(\vec{x},T-t)}\right)\tilde{p}(\vec{x},t)\nonumber\\
&&+\frac{D}{2}g^2(T-t)\nabla^2\tilde{p}(\vec{x},t)+\lambda\int\D\vec{z}\,\rho(z)\left[\tilde{p}(\vec{x}+g(T-t)\vec{z},t)-\tilde{p}(\vec{x},t)\right]\label{genjfpe2}
\ee
using $\tilde{p}(\vec{x},t)=p(\vec{x},T-t)$ and Eq.~(\ref{identity}) in the last step. Eq.~(\ref{genjfpe2}) is the Fokker-Planck equation for the stochastic dynamics
\be
\label{gensde}
\dot{\vec{X}}(t)=-\vec{f}(\vec{X},T-t)+Dg^2(T-t)\nabla \log\,p(\vec{X},T-t)+2\lambda\frac{\blam(\vec{X},T-t)}{p(\vec{X},T-t)}+g(T-t)\bxi(t)
\ee
which likewise represents the generative process.

\subsection{Generalized score function}
\label{Sec:gensc}

In order to determine $\blam$ further, I express $p$ as a Fourier transform and focus on a single component
\be
\label{jdrift}
\Lambda_i(\vec{x},t)&=&g(t)\int_0^1\D \gamma\int\D\vec{z}\,z_i\,\rho(z) \,p(\vec{x}+\gamma g(t)\vec{z},t)\nonumber\\
&=&\frac{g(t)}{(2\pi)^d}\int\D \vec{k}\int_0^1\D \gamma\int\D\vec{z}\,z_i\,\rho(z)\,e^{-i\vec{k}{\cdot}(\vec{x}+\gamma g(t)\vec{z})}\hat{p}(\vec{k},t)\nonumber\\
&=&\frac{1}{(2\pi)^d}\int\D \vec{k}\,e^{-i\vec{k}{\cdot}\vec{x}}\hat{p}(\vec{k},t)\int_0^1\D \gamma\,\frac{i}{\gamma}\frac{\partial}{\partial k_i}\,\phi(k\, \gamma\, g(t)).
\ee
In the last step I used the property
\be
\frac{\partial}{\partial k_i}\,\phi(k\, \gamma\, g(t))&=&\frac{\partial}{\partial k_i}\int\D \vec{z}\left(e^{-i\gamma g(t)\vec{k}{\cdot}\vec{z}}-1\right)\rho(z)\nonumber\\
&=&-i\gamma g(t)\int\D \vec{z}\,z_i\,\rho(z)e^{-i\gamma g(t)\vec{k}{\cdot}\vec{z}}
\ee
due to Eq.~(\ref{phi}) and isotropy. This yields further, reminding that $k=\|\vec{k}\|=\left(\sum_{i=1}^dk_i^2\right)^{1/2}$
\be
\int_0^1\D \gamma\,\frac{i}{\gamma}\frac{\partial}{\partial k_i}\,\phi(k \gamma g(t))&=&\int_0^1\D \gamma\,i g(t)\,\phi'(k\, \gamma\, g(t))\frac{\partial k}{\partial k_i}\nonumber\\
&=&i\,\phi(k g(t))\frac{1}{k}\frac{\partial k}{\partial k_i}\nonumber\\
&=&\frac{i\,k_i}{k^2}\phi(k g(t))
\ee
Substituting back into Eq.~(\ref{jdrift}) yields the result for $\Lambda_i$ as an inverse Fourier transform 
\be
\label{lamresult}
\Lambda_i(\vec{x},t)&=&-\frac{\partial}{\partial x_i}\frac{1}{(2\pi)^d}\int\D \vec{k}\,e^{-i\vec{k}{\cdot}\vec{x}}\hat{p}(\vec{k},t)\frac{\phi(k\, g(t))}{k^2}.
\ee

A notational simplification can be introduced by defining (cf. Eq.~(\ref{jdscore}))
\be
\mathcal{V}(\vec{x},t)&=&\frac{D}{2}g^2(t)\,p(\vec{x},t)-\frac{\lambda}{(2\pi)^d}\int\D \vec{k}\,e^{-i\vec{k}{\cdot}\vec{x}}\hat{p}(\vec{k},t)\frac{\phi(k\, g(t))}{k^2}\nonumber\\
&=&\frac{1}{(2\pi)^d}\int\D \vec{k}\,e^{-i\vec{k}{\cdot}\vec{x}}\hat{p}(\vec{k},t)\frac{\psi(k\, g(t))}{k^2}\label{fdef}
\ee
where $\psi(k)$ is defined in Eq.~(\ref{xicf}). With Eq.~(\ref{lamresult}), we can write the generative process Eq.~(\ref{jpf}) as
\be
\label{jpf2}
\dot{\vec{X}}(t)=-\vec{f}(\vec{X},T-t)+\frac{\nabla\mathcal{V}(\vec{X},T-t)}{p(\vec{X},T-t)}
\ee
Identifying the {\em generalized score function} as in Eq.~(\ref{gensc}) thus leads to Eq.~(\ref{jdpf}). Likewise, Eq.~(\ref{gensde}) leads to Eq.~(\ref{jdsde}).

\section{Denoising score matching loss function}
\label{Sec:loss}

In order to derive the jump-diffusion equivalent of the denoising score matching (DSM) loss function Eq.~(\ref{Gloss}), I start with the explicit score matching (ESM) loss function  for the generalized score function $\mathcal{S}(\vec{x},t)$, Eq.~(\ref{gensc}), and follow the derivation of the conventional DSM loss function in \cite{Vincent:2011aa}:
\be
\mathcal{J}_{\rm ESM}(\btheta)&=&\int_0^T\D t\,w(t)\,\mathbb{E}_{\vec{Y}(t)}\left[\left\|\vec{s}_{\btheta}(\vec{Y},t)-\mathcal{S}(\vec{x},t)\right\|^2\right]\nonumber\\
&=&\int_0^T\D t\,w(t)\,\mathbb{E}_{\vec{Y}(t)}\left[\left\|\vec{s}_{\btheta}(\vec{Y},t)-\frac{\nabla\mathcal{V}(\vec{Y},t)}{p(\vec{Y},t)}\right\|^2\right]\nonumber\\
&=&\int_0^T\D t\,w(t)\,\mathbb{E}_{\vec{Y}(t)}\left[\|\vec{s}_{\btheta}(\vec{Y},t)\|^2-2\vec{s}_{\btheta}(\vec{Y},t){\cdot}\frac{\nabla\mathcal{V}(\vec{Y},t)}{p(\vec{Y},t)}+\left\|\frac{\nabla\mathcal{V}(\vec{Y},t)}{p(\vec{Y},t)}\right\|^2\right]\nonumber\\
&=&\int_0^T\D t\,w(t)\left(\mathbb{E}_{\vec{Y}(t)}\left[\|\vec{s}_{\btheta}(\vec{Y},t)\|^2\right]-2\,\mathbb{E}_{\vec{Y}(t)}\left[\vec{s}_{\btheta}(\vec{Y},t){\cdot}\frac{\nabla\mathcal{V}(\vec{Y},t)}{p(\vec{Y},t)}\right]\right)+C_1(T)\nonumber\\
\label{jesm}
\ee
since the last term is independent of $\btheta$. The function $w(t)$ is a given weight function. For the second term I obtain further
\be
\mathbb{E}_{\vec{Y}(t)}\left[\vec{s}_{\btheta}(\vec{Y},t){\cdot}\frac{\nabla\mathcal{V}(\vec{Y},t)}{p(\vec{Y},t)}\right]&=&\int\D\vec{x}\,\vec{s}_{\btheta}(\vec{x},t){\cdot}\frac{\nabla\mathcal{V}(\vec{x},t)}{p(\vec{x},t)}p(\vec{x},t)\nonumber\\
&=&\int\D\vec{x}\,\vec{s}_{\btheta}(\vec{x},t){\cdot}\nabla\mathcal{V}(\vec{x},t)\nonumber\\
&=&\int\D\vec{x}\,\vec{s}_{\btheta}(\vec{x},t){\cdot}\nabla\int\D\vec{x}'\mathcal{V}(\vec{x},t|\vec{x}')p_{\rm data}(\vec{x}')\label{dsmderiv}
\ee
In analogy to Eq.~(\ref{fdef}) I define now (cf. Eq.~(\ref{jdscorecond}))
\be
\mathcal{V}(\vec{x},t|\vec{x}')&=&\frac{1}{(2\pi)^d}\int\D \vec{k}\,e^{-i\vec{k}{\cdot}\vec{x}}\hat{p}(\vec{k},t|\vec{x}')\frac{\psi(k\, g(t))}{k^2}
\ee
where $\hat{p}(\vec{k},t|\vec{x}')$ denotes the FT of the conditional PDF $p(\vec{x},t|\vec{x}')$. Continuing from Eq.~(\ref{dsmderiv}) yields
\be
\mathbb{E}_{\vec{Y}(t)}\left[\vec{s}_{\btheta}(\vec{Y},t){\cdot}\frac{\nabla\mathcal{V}(\vec{Y},t)}{p(\vec{Y},t)}\right]&=&\int\D\vec{x}\int\D\vec{x}'\,\vec{s}_{\btheta}(\vec{x},t){\cdot}\nabla\mathcal{V}(\vec{x},t|\vec{x}')\frac{p(\vec{x},t|\vec{x}')}{p(\vec{x},t|\vec{x}')}p_{\rm data}(\vec{x}')\nonumber\\
&=&\int\D\vec{x}\int\D\vec{x}'\,\vec{s}_{\btheta}(\vec{x},t){\cdot}\frac{\nabla\mathcal{V}(\vec{x},t|\vec{x}')}{p(\vec{x},t|\vec{x}')}p(\vec{x},t|\vec{x}')p_{\rm data}(\vec{x}')\nonumber\\
&=&\mathbb{E}_{\vec{Y}(0)}\mathbb{E}_{\vec{Y}(t)|\vec{Y}(0)}\left[\vec{s}_{\btheta}(\vec{Y},t){\cdot}\frac{\nabla\mathcal{V}(\vec{Y},t|\vec{Y}_0)}{p(\vec{Y},t|\vec{Y}_0)}\right]\label{jequiv}
\ee
The equivalent of Eq.~(\ref{Gloss}) is thus $\mathcal{J}_{\rm JD}$ of Eq.~(\ref{jdloss}), which satisfies
\be
\mathcal{J}_{\rm JD}(\btheta)&=&\int_0^T\D t\,w(t)\left(\mathbb{E}_{\vec{Y}(t)}\left[\|\vec{s}_{\btheta}(\vec{Y},t)\|^2\right]\right.\nonumber\\
&&\left.-2\,\mathbb{E}_{\vec{Y}(0)}\mathbb{E}_{\vec{Y}(t)|\vec{Y}(0)}\left[\vec{s}_{\btheta}(\vec{Y},t){\cdot}\frac{\nabla\mathcal{V}(\vec{Y},t|\vec{Y}_0)}{p(\vec{Y},t|\vec{Y}_0)}\right]\right)+C_2(T)\nonumber\\
&=&\mathcal{J}_{\rm ESM}(\btheta)-C_1(T)+C_2(T)
\ee
due to Eqs.~(\ref{jesm},\ref{jequiv}) and thus optimizing $\mathcal{J}_{\rm JD}(\btheta)$ with respect to $\btheta$ is equivalent to optimizing $\mathcal{J}_{\rm ESM}(\btheta)$.

\section{The generalized Ornstein-Uhlenbeck process}
\label{Sec:genOU}

{
The forward process under consideration in the detailed implementation is a generalized Ornstein-Uhlenbeck process defined by Eqs.~(\ref{forward},\ref{jnoise}) with
\be
\label{oufg}
\vec{f}(\vec{x},t)=-\frac{1}{2}\,\vec{x}\quad,\qquad\qquad g(t)=1
\ee
I first consider an arbitrary jump amplitude distribution $\rho$ with corresponding $\phi(k)$, Eq.~(\ref{phi}). The Fokker-Planck equation associated with Eqs.~(\ref{forward},\ref{jnoise}) is given in Eq.~(\ref{jfpe}), which we want to solve for the conditional PDF $p(\vec{x},t|\vec{x}')$ with the initial condition $p(\vec{x},0|\vec{x}')=\delta(\vec{x}-\vec{x}')$. Fourier-transforming this Fokker-Planck equation for the choice of $\vec{f},g$ in Eq.~(\ref{oufg}) yields
\be
\frac{\partial}{\partial t}\hat{p}(\vec{k},t|\vec{x}')&=&\left[-\frac{1}{2}\vec{k}{\cdot}\nabla_{\vec{k}} -\frac{D}{2}k^2+\lambda\phi(k)\right]\hat{p}(\vec{k},t|\vec{x}')\nonumber\\
&=&\left[-\frac{1}{2}\vec{k}{\cdot}\nabla_{\vec{k}} -\psi(k)\right]\hat{p}(\vec{k},t|\vec{x}')
\ee
with initial condition $\hat{p}(\vec{k},0|\vec{x}')=e^{i\vec{k}{\cdot}\vec{x}'}$. This first-order equation can be solved with the method of characteristics, which yields the solution
\be
\hat{p}(\vec{k},t|\vec{x}')&=&\exp\left\{i\vec{k}{\cdot}\vec{x}'e^{-t/2}-\int_0^t\D u\,\psi(k\,e^{-u/2})\right\}\nonumber\\
&=&e^{i\vec{k}{\cdot}\vec{x}'e^{-t/2}}\hat{G}(k,t)\label{OUphat}
\ee
where
\be
\label{G1hat}
\hat{G}(k,t)&=&\exp\left\{-\int_0^t\D u\,\psi(k\,e^{-u/2})\right\}\\
&=&\exp\left\{-\frac{D}{2}k^2\left(1-e^{-t}\right)+\lambda\int_0^t\D u\,\phi(k\,e^{-u/2})\right\}
\ee
The conditional PDF is given as the inverse FT of $\hat{p}(\vec{k},t|\vec{x}')$ which can be written in the form
\be
\label{OUcondG1}
p(\vec{x},t|\vec{x}')&=&G(\vec{x}-\vec{x}'e^{-t/2},t)
\ee
with the isotropic function
\be
\label{G1ft}
G(\vec{x},t)&=&\frac{1}{(2\pi)^d}\int\D \vec{k}\,e^{-i\vec{k}{\cdot}\vec{x}}\hat{G}(k,t).
\ee

The definition Eq.(\ref{jdscorecond}) implies that
\be
\label{nablaVk}
\mathcal{F}\left\{\frac{\partial}{\partial x_i} \mathcal{V}(\vec{x},t|\vec{x}')\right\}=-i\,k_i\,e^{i\vec{k}{\cdot}\vec{x}'e^{-t/2}}\hat{G}(k,t)\frac{\psi(k)}{k^2}
\ee
Moreover, Eq.~(\ref{G1hat}) leads to the exact relationship
\be
\label{G2exact}
\frac{\partial}{\partial k_i}\hat{G}(k,t)=2\frac{k_i}{k^2}\left(\psi(k\,e^{-t/2})-\psi(k)\right)\hat{G}(k,t)
\ee
and thus
\be
\label{G2exact2}
k_i\,\hat{G}(k,t)\frac{\psi(k)}{k^2}=-\frac{1}{2}\frac{\partial}{\partial k_i}\hat{G}(k,t)+k_i\,\hat{G}(k,t)\frac{\psi(k\,e^{-t/2})}{k^2}
\ee
While not much has been gained here, the last equation simplifies further for Laplace distributed amplitudes.

\subsection{Laplace distributed amplitudes}

In the following I set $\lambda=1$ and focus on jumps according to an isotropic multivariate Laplace distribution $\text{L}_d(\sigma^2)$, which has the characteristic function Eq.~(\ref{MGL2}) (see Appendix~\ref{Sec:laplace}). The function $\phi$ of Eq.~(\ref{phi}) becomes then
\be
\label{phiL}
\phi_{\rm L}(k)=\frac{1}{1+\frac{\sigma^2}{2}k^2}-1
\ee
and
\be
\int_0^t\D u\,\phi_{\rm L}(k\,e^{-u/2})=\log\left(\frac{1+\frac{\sigma^2}{2}k^2e^{-t}}{1+\frac{\sigma^2}{2}k^2}\right)
\ee
As a consequence, $\hat{G}(k,t)$ of Eq.~(\ref{G1hat}) assumes the form
\be
\label{G1sol}
\hat{G}(k,t)&=&\left(\frac{1+\frac{\sigma^2}{2}k^2e^{-t}}{1+\frac{\sigma^2}{2}k^2}\right)e^{-\frac{D}{2}k^2\left(1-e^{-t}\right)}\\
&=&\left(\frac{1-e^{-t}}{1+\frac{\sigma^2}{2}k^2}+e^{-t}\right)e^{-\frac{D}{2}k^2\left(1-e^{-t}\right)}\label{G1solb}
\ee

Since $\psi(k)=\frac{D}{2}k^2-\phi_{\rm L}(k)$, we also obtain 
\be
\label{psiratiot}
\frac{\psi(k\,e^{-t/2})}{k^2}=\frac{D}{2}e^{-t}+\frac{\frac{\sigma^2}{2}e^{-t}}{1+\frac{\sigma^2}{2}k^2e^{-t}}
\ee
Multiplying Eq.~(\ref{psiratiot}) with Eq.~(\ref{G1sol}), the last term in Eq.~(\ref{G2exact2}) now becomes
\be
\hat{G}(k,t)\frac{\psi(k\,e^{-t/2})}{k^2}&=&\frac{D}{2}e^{-t}\hat{G}(k,t)+\frac{\frac{\sigma^2}{2}e^{-t}}{1+\frac{\sigma^2}{2}k^2}e^{-\frac{D}{2}k^2\left(1-e^{-t}\right)}\label{G2exact3}\\
&=&\frac{D}{2}e^{-t}\hat{G}(k,t)+\frac{\frac{\sigma^2}{2}e^{-t}}{1-e^{-t}}\left(\hat{G}(k,t)-e^{-t}e^{-\frac{D}{2}k^2\left(1-e^{-t}\right)}\right)\label{G2exact4}
\ee
where in the last step Eq.~(\ref{G1solb}) has been used. Substituting Eq.~(\ref{G2exact4}) in Eq.~(\ref{G2exact2}) and subsequently Eq.~(\ref{nablaVk}) yields
\be
\label{nablaVk2}
\mathcal{F}\left\{\frac{\partial}{\partial x_i} \mathcal{V}(\vec{x},t|\vec{x}')\right\}&=&i\,e^{i\vec{k}{\cdot}\vec{x}'e^{-t/2}}\left(\frac{1}{2}\frac{\partial}{\partial k_i}\hat{G}(k,t)-k_i\left(\frac{D}{2}e^{-t}+\frac{\frac{\sigma^2}{2}e^{-t}}{1-e^{-t}}\right)\hat{G}(k,t)\right.\nonumber\\
&&\left.+\,k_i\,\frac{\frac{\sigma^2}{2}e^{-2t}}{1-e^{-t}}e^{-\frac{D}{2}k^2\left(1-e^{-t}\right)}\right)
\ee
The inverse FT is now exact
\be
\label{nablaVx}
\nabla \mathcal{V}(\vec{x},t|\vec{x}')&=&-\frac{1}{2}\left(\vec{x}-\vec{x}'e^{-t/2}\right)p(\vec{x},t|\vec{x}')+\left(\frac{D}{2}e^{-t}+\frac{\frac{\sigma^2}{2}e^{-t}}{1-e^{-t}}\right)\nabla p(\vec{x},t|\vec{x}')\nonumber\\
&&-\frac{\frac{\sigma^2}{2}e^{-2t}}{1-e^{-t}}\nabla\mathcal{N}(\vec{x}-\vec{x}'e^{-t/2},D(1-e^{-t})\vec{I}_{d})
\ee
where $\mathcal{N}(\boldsymbol\mu,\boldsymbol\Sigma)$ denotes the $d$-dimensional multivariate normal distribution with mean vector $\boldsymbol\mu$ and covariance matrix $\boldsymbol\Sigma$. Unfortunately, $p(\vec{x},t|\vec{x}')$ as inverse FT of Eq.~(\ref{G1solb}) can not be derived in closed analytical form. In order to have an easily applicable generative model, I further consider the pure jump case $D=0$.

\subsection{The JL model}

Setting $D=0$ yields in Eq.~(\ref{G1sol})
\be
\label{G1solD0}
\hat{G}(k,t)&=&\frac{1-e^{-t}}{1+\frac{\sigma^2}{2}k^2}+e^{-t}
\ee
which has the exact inverse FT (see Appendix~\ref{Sec:laplace})
\be
G(\vec{x},t)&=&\rho_{\rm L}(\vec{x},\sigma)\left(1-e^{-t}\right)+\delta(\vec{x})e^{-t}
\ee
where $\rho_{\rm L}(\vec{x},\sigma)$ is the distribution $\text{L}_d(\sigma^2)$ itself, see Eq.~(\ref{rhogl}). The conditional PDF is thus determined with Eq.~(\ref{OUcondG1}).

In addition, setting $D=0$ in Eq.~(\ref{nablaVx}) yields
\be
\label{nablaVxD0}
\nabla \mathcal{V}(\vec{x},t|\vec{x}')&=&-\frac{1}{2}\left(\vec{x}-\vec{x}'e^{-t/2}\right)p(\vec{x},t|\vec{x}')\nonumber\\
&&+\left(\frac{\frac{\sigma^2}{2}e^{-t}}{1-e^{-t}}\right)\nabla\left(p(\vec{x},t|\vec{x}')-\delta\left(\vec{x}-\vec{x}'e^{-t/2}\right)e^{-t}\right)\nonumber\\
&=&-\frac{1}{2}\left(\vec{x}-\vec{x}'e^{-t/2}\right)p(\vec{x},t|\vec{x}')+\frac{\sigma^2}{2}e^{-t}\nabla\rho_{\rm L}(\vec{x}-\vec{x}'e^{-t/2},\sigma)\nonumber\\
&=&-\frac{1}{2}\left(\vec{x}-\vec{x}'e^{-t/2}\right)p(\vec{x},t|\vec{x}')\nonumber\\
&&+\frac{\sigma^{2-d}\,e^{-t}}{(2\pi)^{d/2}}\nabla\left(\frac{\|\vec{x}-\vec{x}'e^{-t/2}\|}{\sqrt{2}\,\sigma}\right)^{1-\frac{d}{2}}K_{1-\frac{d}{2}}\left(\frac{\|\vec{x}-\vec{x}'e^{-t/2}\|}{\sigma/\sqrt{2}}\right)
\ee
After evaluating the derivatives of the Bessel function and simplifications, we obtain the conditional generalized score function in the form
\be
\mathcal{S}(\vec{x},t)=\frac{\nabla\mathcal{V}(\vec{x},t|\vec{x}')}{p(\vec{x},t|\vec{x}')}=\frac{\vec{x}-\vec{x}'e^{-t/2}}{\|\vec{x}-\vec{x}'e^{-t/2}\|}\mathcal{G}(\|\vec{x}-\vec{x}'e^{-t/2}\|,t)
\ee
where $\mathcal{G}(x,t)$ is defined in Eq.~(\ref{G2jl}).

\subsection{Variance of the score estimator}

\begin{figure}
\centering
\includegraphics[width=6cm]{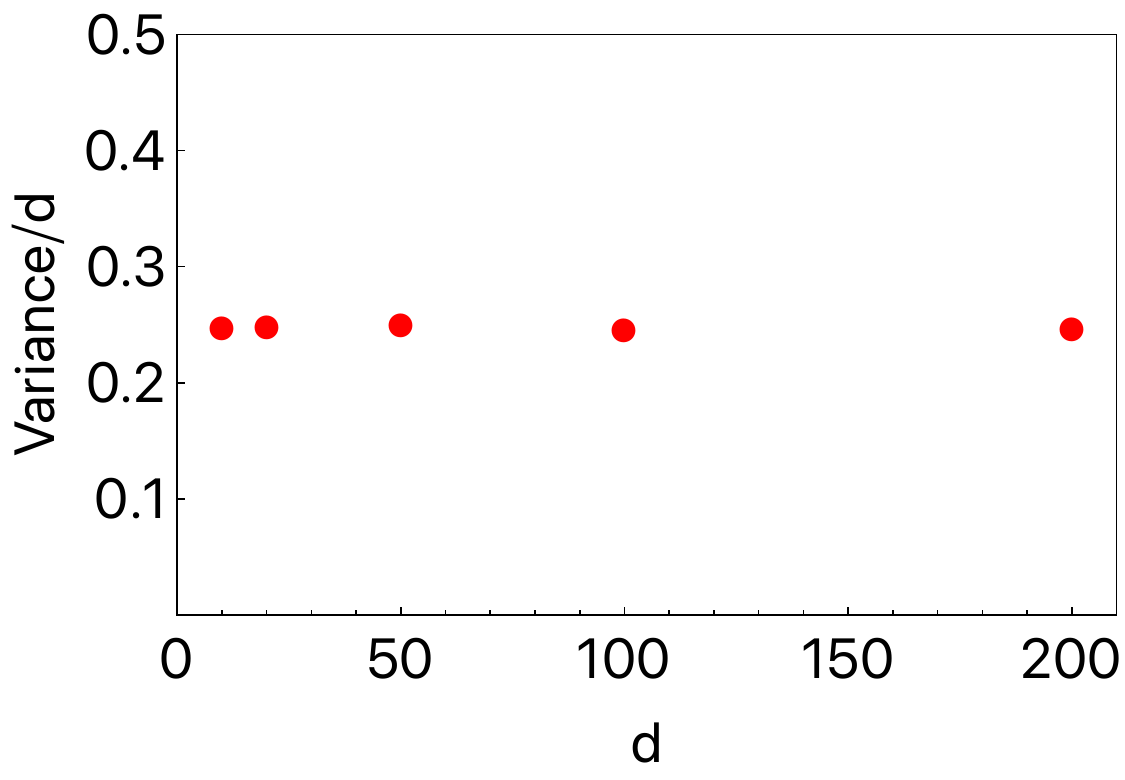}
\caption{\label{Fig:variance}{The lhs of Eq.~\ref{app_var} obtained by Monte-Carlo sampling and divided by $d$. The increments $\vec{J}(t)$ are sampled as outlined in Sec.~\ref{Sec:jl_sampling}.}}
\end{figure}

In the denoising score matching loss function Eq.~(\ref{jdloss_scl2}) one can identify the score estimator $\frac{\vec{J}(t)}{\|\vec{J}(t)\|}\hat{\mathcal{G}}(\|\vec{J}(t)\|,t)$, where $\hat{\mathcal{G}}(x,t)=(1-e^{-t})\mathcal{G}(x,t)$ and the jump increments $\vec{J}(t)$ have the distribution Eq.~(\ref{Jdist}). While the variance of this estimator can not be evaluated analytically, it is straightforward to compute it by sampling. As Fig.~\ref{Fig:variance} shows, for constant $t$ the variance
\be 
\label{app_var}
\sum_{j=1}^d{\rm Var}\left(\frac{J_i(t)}{\|\vec{J}(t)\|}\hat{\mathcal{G}}(\|\vec{J}(t)\|,t))\right)\propto d
\ee
then scales linearly in $d$ just like in the Gaussian case. In this case, the estimator is just the noise itself $\boldsymbol\epsilon\sim \mathcal{N}(\vec{0},\vec{I}_d)$, see Eq.~\ref{Gloss_exp}.
}

\section{The multivariate Laplace distribution}
\label{Sec:laplace}

I briefly review basic properties of the multivariate Laplace distribution following \cite{Kozubowski:2013aa}.

{\bf Definition} {\em (Multivariate generalized Laplace law)}. A random vector in $\mathbb{R}^d$ is said to have a multivariate generalized asymmetric Laplace distribution, if its characteristic function is given by
\be
\label{MGL1}
\chi(\vec{k})=\left(\frac{1}{1+\frac{1}{2}\vec{k}^{\rm T}\bsigma\vec{k}-i\bmu{\cdot}\vec{k}}\right)^\nu,\qquad\qquad \vec{k}\in\mathbb{R}^d
\ee
where $\nu>0$, $\bmu\in\mathbb{R}^d$, and $\bsigma$ is a $d\times d$ non-negative definite symmetric matrix. This distribution is denoted by $\text{GAL}(\bmu,\bsigma,\nu)$.

Two relevant properties:
\begin{enumerate}
\item A random vector $\vec{L}\sim \text{GAL}(\bmu,\bsigma,\nu)$ can be interpreted as a subordinated Gaussian process such that
\be
\label{GLsampling}
\vec{L}\overset{d}{=}\bmu \Gamma+\sqrt{\Gamma}\,\vec{X}
\ee
where $\vec{X}\sim \mathcal{N}(\vec{0},\bsigma)$ and $\Gamma$ has a standard Gamma distribution with shape parameter $\nu$.

\item The mean vector and covariance matrix of $\vec{L}\sim \text{GAL}(\bmu,\bsigma,\nu)$ are given by
\be
\mathbb{E}(\vec{L})=\bmu\,\nu,\qquad\qquad \text{Cov}(\vec{L})=\nu(\bsigma+\bmu{\cdot}\bmu)
\ee

\end{enumerate}

Since we are interested in describing the jump amplitudes in the noise process $\bxi(t)$, Eq.~(\ref{jnoise}), we want no contribution to the drift leading to the constraint $\bmu=\vec{0}$. We also require an isotropic distribution, which is realized by setting $\bsigma=\sigma^2\vec{I}$.

For the JL model only the $\nu=1$ value is relevant leading to the {\em isotropic multivariate generalized Laplace distribution} $\text{L}_d(\sigma^2)=\text{GAL}(\vec{0},\sigma^2\vec{I},1)$, which has the characteristic function
\be
\label{MGL2}
\chi_{\rm L}(\vec{k})=\left(1+\frac{\sigma^2}{2}k^2\right)^{-1}
\ee
Amplitude samples can be easily generated using Eq.~(\ref{GLsampling}): $\vec{L}\sim\text{L}_d(\sigma^2)$ has the representation
\be
\label{GLsampling2}
\vec{L}\overset{d}{=}\sqrt{\Gamma}\,\vec{X}
\ee
where $\vec{X}\sim \mathcal{N}(\vec{0},\sigma^2\vec{I})$ and $\Gamma\sim \text{Gamma}(1)$.

Note that the PDFs associated with the characteristic function Eqs.~(\ref{MGL1},\ref{MGL2}) have closed analytical forms \cite{Kozubowski:2013aa}. For Eq.~(\ref{MGL2}) one obtains the isotropic jump PDF
\be
\label{rhogl}
\rho_{\rm L}(\vec{z},\sigma)&=&\mathcal{F}^{-1}\left\{\left(1+\frac{\sigma^2}{2}k^2\right)^{-1}\right\}\nonumber\\
&=&\frac{2}{\sigma^d(2\pi)^{d/2}}\left(\frac{z}{\sigma \sqrt{2}}\right)^{1-\frac{d}{2}}K_{1-\frac{d}{2}}\left(\frac{z}{\sigma}\sqrt{2}\right)\label{MGLFT}
\ee
where $K_\mu$ denotes the modified Bessel function of the 2nd kind.

\section{Distribution of the jump increments $\tilde{\vec{J}}(t)$}
\label{Sec:jtilde}

{From the definition Eq.~(\ref{incrdef2}) we can directly derive the characteristic function $\chi(\vec{k},t)$ of $\tilde{\vec{J}}(t)$ by evaluating the expected value:
\be
\chi(\vec{k},t)=\mathbb{E}\left[\exp\left\{i\vec{k}\sum_{j=1}^{N_t}\vec{A}_je^{(t-\tau_j)/2}\right\}\right]
\ee
where $N_t\sim \text{Poisson}(\lambda t)$, $\tau_j\sim \text{U}([0,t])$, and $\vec{A}_j\sim\text{L}_d(\sigma^2)$ i.i.d.. Calculating the expected value over $N_t,\tau$ is straightforward 
\be
\chi(\vec{k},t)&=&\exp\left\{\lambda\int_0^t\D s\left(\mathbb{E}\left[e^{i\vec{k}\vec{A}e^{(t-s)/2}}\right]-1\right)\right\}\nonumber\\
&=&\exp\left\{\lambda\int_0^t\D s\,\phi_{\rm L}\left(k\,e^{s/2}\right)\right\}
\ee
using Eq.~(\ref{phi}). Direct integration with Eq.~(\ref{phiL}) yields
\be
\int_0^t\D s\,\phi_{\rm L}\left(k\,e^{s/2}\right)=-\log\left(\frac{1+\frac{\sigma^2}{2}k^2e^{t}}{1+\frac{\sigma^2}{2}k^2}\right)
\ee
The characteristic function is then
\be
\chi(\vec{k},t)&=&\frac{1+\frac{\sigma^2}{2}k^2}{1+\frac{\sigma^2}{2}k^2e^{t}}\nonumber\\
&=&\frac{1}{1+\frac{\sigma^2}{2}k^2e^{t}}(1-e^{-t})+e^{-t}
\ee
As a consequence, the PDF of $\tilde{\vec{J}}(t)$ is given by
\be
p(\tilde{\vec{J}},t)=\tilde{\rho}_{\rm L}(\|\tilde{\vec{J}}\|,\sigma\,e^{t/2})(1-e^{-t})+\delta(\tilde{\vec{J}})e^{-t}
\ee
and can be sampled as outlined in Sec.~\ref{Sec:jl_sampling}.}

\section{Details on the numerical experiments}
\label{Sec:exp}

{Two data distributions are considered: (a) a swiss roll distribution standardized to zero mean and unit variance; (b) a mixture of 9 equal Gaussians with variance $\sigma^2=0.05^2$ arranged in a grid with weights
\be
\vec{W}=\left(\begin{matrix} 0.3 & 0.15 & 0.05 \\ 0.1 & 0.02 & 0.15 \\ 0.01 & 0.2 & 0.02 \end{matrix}\right)
\ee
The Gaussian mixture distribution (b) is adapted from \cite{Shariatian:2025aa}.

Score functions are estimated by minimizing the DSM loss functions Eq.~(\ref{jdloss_scl2}) (JL-ODE/SDE), Eq.~(\ref{Gloss_exp}) (Gauss-ODE), and Eq.~(\ref{LIMloss}) (LIM-SDE) using $10^5$ training samples. The neural network architecture is a simple multilayer perceptron with 4 layers of 124 nodes and GELU activation functions. Training is performed using an ADAM optimizer, a learning rate of $\approx 10^{-3}$, and a batch size of 64. All code is implemented in {\bf Mathematica}.

\subsection{The Gaussian probability flow ODE}

I follow the standard formalism \cite{Song:2021aa} with implementation improvements discussed in \cite{Zhang:2023aa}. The forward process is given as an OU process with $\vec{f}(\vec{x},t)=\frac{1}{2}\vec{x}$ and $g(t)=1$ in Eq.~(\ref{forward}) and $\bxi$ Gaussian white noise. Rescaling the standard score function in Eq.~(\ref{Gloss}) then yields the loss function
\begin{equation}
\label{Gloss_exp}
J_{\rm G}(\btheta)=\mathbb{E}_{t\sim{\rm U}(0,T)}\mathbb{E}_{\vec{Y}_0\sim p_{\rm data}}\mathbb{E}_{\boldsymbol\epsilon\sim\mathcal{N}(\vec{0},\vec{I})}\left[\left\|\hat{\vec{s}}_{\btheta}\left(\vec{Y}_0e^{-t/2}+\sqrt{D(1-e^{-t})}\,\boldsymbol\epsilon,t\right)-\boldsymbol\epsilon\right\|^2\right]
\end{equation}
implying a weight function $w(t)=\left(1-e^{-t}\right)/T$. The generative process is given by the probability flow ODE in exponential integrator form \cite{Zhang:2023aa}
\be
\vec{X}_{i+1}=\vec{X}_i\,e^{\Delta t/2}-\left(e^{\Delta t/2}\sqrt{1-e^{-(T-t)}}-\sqrt{1-e^{\Delta t-(T-t)}}\right)\sqrt{D}\,\hat{\vec{s}}_{\btheta}(\vec{X}_i,T-t)
\ee
and the starting point is distributed as $\vec{X}_0\sim\mathcal{N}(\vec{0},D\vec{I})$.

\subsection{The LIM SDE} 

As a variant of LIM \cite{Yoon:2023aa} I consider the forward process as a generalized OU process driven by isotropic $\alpha$-stable noise with $1< \alpha\le 2$ 
\be
\label{mylim}
\dot{\vec{Y}}(t)=-\frac{1}{\alpha}\,\vec{Y}(t)+\left(\frac{\sigma^2}{2}\right)^{1/\alpha}\bxi_\alpha(t)
\ee
where the noise increments $\Delta \bxi_\alpha\sim S\alpha S(\Delta t)$ and the isotropic $\alpha$-stable distribution $S\alpha S(\nu)$ is described by the characteristic function
\be
\chi_\alpha(\vec{k})=e^{-\nu\,k^\alpha}
\ee
The factor $\left(\frac{\sigma^2}{2}\right)^{1/\alpha}$ is introduced as a scale parameter such that in the Gaussian limit $\alpha=2$ the noise intensity is given by $\sigma^2$. 
As shown in \cite{Yoon:2023aa} a loss function for the score function estimation is given by
\begin{equation}
\label{LIMloss}
J_{\rm LIM}(\btheta)=\mathbb{E}_{t\sim{\rm U}(0,T)}\mathbb{E}_{\vec{Y}_0\sim p_{\rm data}}\mathbb{E}_{\boldsymbol\epsilon\sim S\alpha S(1)}\left[\left\|\hat{\vec{s}}_{\btheta}\left(\vec{Y}_0e^{-t/\alpha}+\gamma(t)\,\boldsymbol\epsilon,t\right)-\boldsymbol\epsilon\right\|^2\right]
\end{equation}
where $\gamma(t)=\left(\frac{\sigma^2}{2}\left(1-e^{-t}\right)\right)^{1/\alpha}$. The SDE of the generative process follows in the exponential integrator form
\be
\vec{X}_{i+1}=\vec{X}_i\,e^{\Delta t/\alpha}-\alpha\,\frac{\sigma^2\left(e^{\Delta t/\alpha}-1\right)}{2\gamma(T-t)^{\alpha-1}}\,\hat{\vec{s}}_{\btheta}(\vec{X}_i,T-t)+\left(\frac{\sigma^2}{2}\left(e^{\Delta t}-1\right)\right)^{1/\alpha}\boldsymbol\epsilon
\ee
where $\boldsymbol\epsilon\sim S\alpha S(1)$ and the starting point is distributed as $\vec{X}_0\sim S\alpha S\left(\frac{\sigma^2}{2}\right)$.

}

\section{Alternative derivation of the continuity equation for $\tilde{p}$}
\label{Sec:alt}

The time-reversal of Markovian jump processes has been rigorously investigated in \cite{Conforti:2022aa}. Considering the case $D=0$, the generator of the time-reversal of $\vec{Y}(t)$, Eqs.~(\ref{forward},\ref{xicf},\ref{phi}), is shown to have the form
\be
\label{exactrev}
\tilde{\mathcal{L}}u(\vec{x})&=&-\vec{f}(\vec{x},T-t){\cdot}\nabla u(\vec{x})\nonumber\\
&&+\lambda\int\D\vec{z}\,\rho(z)\left[u(\vec{x}+g(T-t)\vec{z})-u(\vec{x})\right]\frac{p(\vec{x}+g(T-t)\vec{z},T-t)}{p(\vec{x},T-t)}
\ee
where $p(\vec{x},t)$ is the PDF of $\vec{Y}(t)$. Using the same identity as in Eq.~(\ref{identity}) leads to
\be
\tilde{\mathcal{L}}u(\vec{x})&=&-\vec{f}(\vec{x},T-t){\cdot}\nabla u(\vec{x})+\lambda\, g(T-t)\int\D\vec{z}\,\rho(z)\int_0^1\D \gamma\, \left[\vec{z}{\cdot}\nabla u(\vec{x}+\gamma g(T-t)\vec{z})\right]\times\nonumber\\
&&\frac{p(\vec{x}+g(T-t)\vec{z},T-t)}{p(\vec{x},T-t)}
\ee
This equation can be used to determine the adjoint generator $\tilde{\mathcal{L}}^\dagger$ that governs the time evolution of $\tilde{p}(\vec{x},t)$, the PDF of the time-reversed process $\vec{X}(t)$:
\be
\int\D \vec{x}\,\tilde{p}(\vec{x},t)\tilde{\mathcal{L}}u(\vec{x}) &=&-\int\D \vec{x}\,\tilde{p}(\vec{x},t)\vec{f}(\vec{x},T-t){\cdot}\nabla u(\vec{x})+\lambda\, g(T-t)\int\D \vec{x}\,\tilde{p}(\vec{x},t)\times\nonumber\\
&&\int_0^1\D \gamma\int\D\vec{z}\,\rho(z)\, \left[\vec{z}{\cdot}\nabla u(\vec{x}+\gamma g(T-t)\vec{z})\right]\frac{p(\vec{x}+g(T-t)\vec{z},T-t)}{p(\vec{x},T-t)}\nonumber\\
&=&\int\D \vec{x}\,u(\vec{x})\nabla{\cdot}\vec{f}(\vec{x},T-t)\tilde{p}(\vec{x},t)+\lambda\, g(T-t)\int\D \vec{x}\int_0^1\D \gamma\int\D\vec{z}\times\nonumber\\
&& \left[\vec{z}{\cdot}\nabla u(\vec{x}+\gamma g(T-t)\vec{z})\right]p(\vec{x}+g(T-t)\vec{z},T-t)\nonumber\\
&=&\int\D \vec{x}\,u(\vec{x})\nabla{\cdot}\vec{f}(\vec{x},T-t)\tilde{p}(\vec{x},t)-\lambda\, g(T-t)\int\D \vec{x}\,u(\vec{x})\times\nonumber\\
&&\int_0^1\D \gamma\int\D\vec{z}\,\rho(z)\,\vec{z}{\cdot}\nabla p(\vec{x}+(1-\gamma)g(T-t)\vec{z},T-t)\nonumber\\
&=&\int\D \vec{x}\,u(\vec{x})\nabla{\cdot}\vec{f}(\vec{x},T-t)\tilde{p}(\vec{x},t)-\lambda\, g(T-t)\int\D \vec{x}\,u(\vec{x})\times\nonumber\\
&&\int_0^1\D \gamma\int\D\vec{z}\,\rho(z)\,\vec{z}{\cdot}\nabla\, \frac{p(\vec{x}+\gamma g(T-t)\vec{z},T-t)}{p(\vec{x},T-t)}\tilde{p}(\vec{x},t)
\ee
using $\tilde{p}(\vec{x},t)=p(\vec{x},T-t)$. Thus the adjoint generator is given by
\be
\tilde{\mathcal{L}}^\dagger \tilde{p}(\vec{x},t)&=&-\nabla{\cdot}\left(-\vec{f}(\vec{x},T-t)+\lambda\, g(T-t)\int_0^1\D \gamma\int\D\vec{z}\,\vec{z}\,\rho(z)\right.\times\nonumber\\
&& \left.\frac{p(\vec{x}+\gamma g(T-t)\vec{z},T-t)}{p(\vec{x},T-t)}\right)\tilde{p}(\vec{x},t).
\ee
In the presence of the additional Gaussian noise component in Eq.~(\ref{xicf}) for $D\neq 0$ one simply adds the standard diffusive term of Eq.~(\ref{deriv1}). This yields Eq.~(\ref{genjfpe}) for the time-evolution of $\tilde{p}$.


\end{document}